\documentclass[lettersize,journal]{IEEEtran}
\usepackage{amsmath,amsfonts}
\usepackage{algorithmic}
\usepackage{algorithm}
\usepackage{array}
\usepackage[caption=false,font=normalsize,labelfont=sf,textfont=sf]{subfig}
\usepackage{textcomp}
\usepackage{stfloats}
\usepackage{amssymb,amsmath}
\usepackage{url}
\usepackage{hyperref}
\usepackage{multirow}
\usepackage{verbatim}
\usepackage{graphicx}
\usepackage{cite}
\usepackage{cancel}
\usepackage[table]{xcolor}
\newcommand{\yzl}[1]{#1}

\begin{document}

\title{DVP-MVS++: Synergize Depth-Normal-Edge and Harmonized Visibility Prior for Multi-View Stereo}

\author{Zhenlong Yuan*, Dapeng Zhang*, Zehao Li, Chengxuan Qian, Jianing Chen, Yinda Chen, Kehua Chen, Tianlu Mao, Zhaoxin Li, Hao Jiang and Zhaoqi Wang~\IEEEmembership{Member,~IEEE,}
\thanks{\\
$^{*}$~The authors contributed equally to this work. 

This work This work was supported in part by the Science and Technology Projects of the Ministry of Agriculture and Rural Affairs of China, the Strategic Priority Research Program of the Chinese Academy of Sciences under Grant No. XDB1290302, and the National Natural Science Foundation of China under Grant 62172392.

Zhenlong Yuan, Zehao Li, Jianing Chen, Kehua Chen, Tianlu Mao, Hao Jiang and Zhaoqi Wang are with the Institute of Computing Technology, Chinese Academy of Sciences, Beijing 100190, China, and also with University of Chinese Academy of Sciences, Beijing, 100049, China. (e-mail: yuanzhenlong21b@ict.ac.cn, lizehao23z@ict.ac.cn, chenjianing23s@ict.ac.cn, chenkehua23s@ict.ac.cn, ltm@ict.ac.cn, jianghao@ict.ac.cn, zqwang@ict.ac.cn)  

Dapeng Zhang is with DSLAB, School of Information Science \& Engineering, Lanzhou University, 730000, China. (zhangdp22@lzu.edu.cn)  

Chengxuan Qian is with School of Mathematical Science, Jiangsu University, 212013, China. (chengxuan.qian@stmail.ujs.edu.cn)  

Yinda Chen is with MoE Key Laboratory of Brain-inspired Intelligent Perception and Cognition, University of Science and Technology of China, Hefei 230027, China (e-mail: cyd0806@mail.ustc.edu.cn).  

Zhaoxin Li is with Agricultural Information Institute, Chinese Academy of Agricultural Sciences, Beijing, 100081, China, and also with Key Laboratory of Agricultural Big Data, Ministry of Agriculture and Rural Affairs, Beijing, 100081, China. (e-mail: cszli@hotmail.com)
}

}

\markboth{}%
{Shell \MakeLowercase{\textit{et al.}}: A Sample Article Using IEEEtran.cls for IEEE Journals}

\maketitle

\begin{abstract}
Recently, patch deformation-based methods have demonstrated significant effectiveness in multi-view stereo due to their incorporation of deformable and expandable perception for reconstructing textureless areas. 
However, these methods generally focus on identifying reliable pixel correlations to mitigate matching ambiguity of patch deformation, while neglecting the deformation instability caused by edge-skipping and visibility occlusions, which may cause potential estimation deviations. 
To address these issues, we propose DVP-MVS++, an innovative approach that synergizes both depth-normal-edge aligned and harmonized cross-view priors for robust and visibility-aware patch deformation. Specifically, to avoid edge-skipping, we first apply DepthPro, Metric3Dv2 and Roberts operator to generate coarse depth maps, normal maps and edge maps, respectively. These maps are then aligned via an erosion-dilation strategy to produce fine-grained homogeneous boundaries for facilitating robust patch deformation. 
\yzl{Moreover, we reformulate view selection weights as visibility maps, and then implement both an enhanced cross-view depth reprojection and an area-maximization strategy to help reliably restore visible areas and effectively balance deformed patch.} 
\yzl{Additionally, we obtain geometry consistency by adopting both aggregated normals via view selection and projection depth differences via epipolar lines, and then employ SHIQ for highlight correction to facilitate highlight perception capacity, thus improving reconstruction quality during propagation and refinement stage. }
Evaluations on ETH3D, Tanks \& Temples and Strecha datasets exhibit the state-of-the-art performance and robust generalization capability of our proposed method.

\end{abstract}


\section{Introduction}
\IEEEPARstart{M}{ulti-view} Stereo (MVS) is a core task in computer vision that aims to densify the geometric representation of a scene or object using overlapping images captured from different viewpoints.
Its wide application spans across augmented reality \cite{Augmented}, autonomous driving \cite{Autonomous}, 3D printing \cite{3D-printed}, etc.
\yzl{In recent years, the emergence of numerous innovative ideas \cite{SED-MVS, MSP-MVS, wang2024efficient, fang2020ugnet, chen2024dsc3d, li2024focus, ye2024self} has significantly boosted its performance across multiple benchmarks \cite{ETH3D, TNT, strecha, Blendedmvs}. These advancements can be broadly categorized into learning-based and traditional MVS.}

\begin{figure}
    \centering \includegraphics[width=\linewidth]{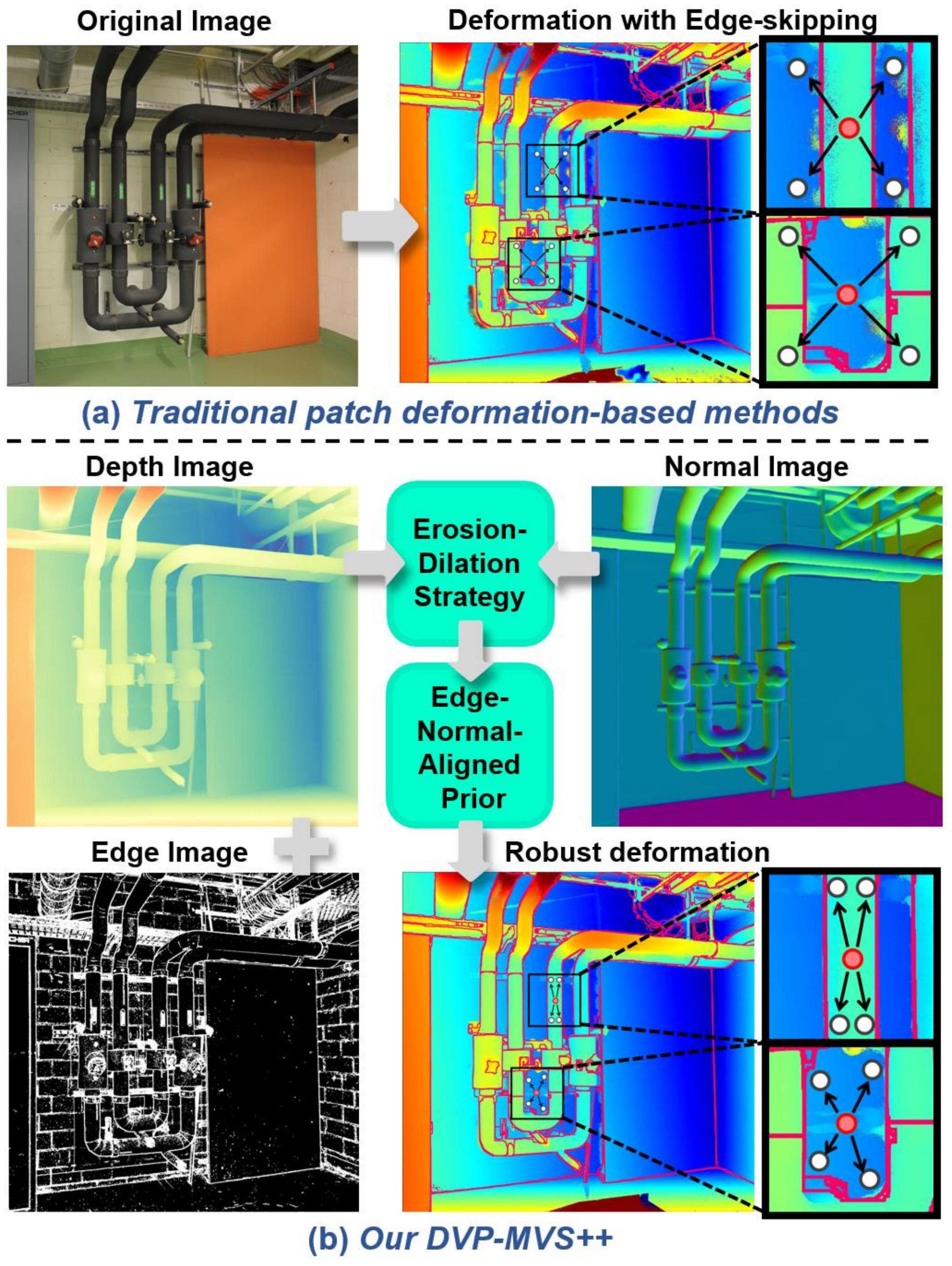}
    \caption{
    Comparison between traditional patch deformation-based methods and our DVP-MVS++. As can be seen, other methods (a) erroneously cross the depth edges to select depth-discontinuous gray pixels as anchors for patch deformation of the central red pixel. Differently, our DVP-MVS++ (b) subtly combines monocular depth maps, normal maps and edge maps to obtain depth-normal-edge aligned prior as guidance to ensure gray pixels being selected in homogeneous areas of red pixel, thus achieving robust patch deformation. 
    }
    \label{fig1}
\end{figure}

Learning-based MVS methods utilize convolutional neural networks to extract high-dimensional cost volumes for reconstruction. However, most of them require large memory usage and tend to have poor generalization performance.
In contrast, traditional MVS methods, which are derived from the PatchMatch algorithm \cite{PM}, propagate potential neighbor hypotheses and randomly generate new hypotheses to construct the solution space, then take the multi-view matching cost as the criterion to selects the optimal one from them. 
However, in textureless areas, the matching cost might becomes unreliable due to the absence of reliable features within its receptive field.

For the reconstruction of textureless areas, two main types of traditional MVS are proposed: planarization-based and patch deformation-based methods. Planarization-based methods link reliable pixels in well-textured areas to form boundaries that separate textureless areas, and then planarize these areas for reconstruction. 
Numerous strategies concerning both area recognition and planarization have been proposed including superpixel segmentation \cite{TSAR-MVS}, triangulation \cite{ACMMP}, and KD-tree \cite{HPM-MVS}.
However, these approaches are typically constrained by the limited size of connected regions, making them susceptible to errors after the planarization process.

Differently, patch deformation-based methods apply deformation on patches to make sure its receptive field contains sufficient features for matching costs. 
For instance, PHI-MVS \cite{PHI-MVS} subtly adopts dilated convolution on patches to dynamically increase patch size and sampling intervals. Differently, APD-MVS \cite{APD-MVS} adaptively searches for correlated reliable pixels around each unreliable pixel, thereby constructing multiple sub-patches centered on them as deformed patch for matching costs. 
\yzl{However, these algorithms mainly focus on exploring advanced strategies for searching reliable pixels to reduce matching ambiguity, while ignore the \textbf{unexpected edge-skipping}, which may cause \textbf{deformation instability.}
Specifically, as shown in Fig.~\ref{fig1}(a), traditional patch deformation-based methods reconstruct \textbf{unreliable} red pixels (i.e., pixels with matching ambiguity) by dividing their surrounding areas into multiple fixed-angle sectors. Within each sector, a reliable gray pixel is identified as the sub-patch center. Then all sub-patches forms the deformed patch for reconstruction. 
Nonetheless, the search process may experience unintended edge-skipping due to shadows or occlusions, which violates depth continuity and leads to \textbf{deformation instability.}}

Therefore, we synergize \textbf{D}epth-Normal-Edge and Harmonized \textbf{V}isibility \textbf{P}rior for \textbf{M}ulti-\textbf{V}iew \textbf{S}tereo (DVP-MVS++).
Specifically, we first leverage DepthPro\cite{depany2}, Metric3Dv2\cite{Metric3D}, and Roberts operator to generate monocular depth maps, normal maps and edge maps, respectively. 
The depth maps provide monocular depth information for \textbf{homogeneous areas} (i.e., areas with depth continuity) in the global level while lack detailed edge information. 
Differently, the normal maps offers detailed geometric structure, making it suitable for fine-grained edge extraction and alignment.
While the edge maps accurately captures rough edges and fragmented regions. 
\yzl{Consequently, we use erosion and dilation techniques to merge the depth, normal, and edge maps. This process creates depth boundaries as \textbf{depth-normal-edge prior} to ensure patches deform only within regions of consistent depth, as shown in Fig.~\ref{fig1}(b).
}

Moreover, visibility occlusion presents another critical challenge for unstable patch deformation.
To address this, we first construct visibility maps through the reformulation of view selection weights~\cite{ACMMP}.
\yzl{Then we convert view selection weights into visibility maps. Using cross-view depth reprojection and area-maximization strategy to balance patch coverage. Regarding the whole process as \textbf{the cross-view prior}, we effectively restore visible regions and avoid overlapping patches.}

\yzl{Finally, we aggregate hemispherical to constrain normal range and apply depth projection to restrict depth intervals, thus obtaining the geometric consistency for both propagation and refinement. Building on this, we further apply SHIQ to enable patch adaptively perceive highlights, thereby achieving \textbf{highlight correction capacity.}}

Compared to the prior work~\cite{DVP-MVS},we further adopt monocular normals for fine-grained depth edge extraction and alignment, denoted as DVP-MVS++, which significantly improves reconstruction quality. The differences are as follows: 
1) In this work, we integrate both monocular depth and normal estimation with edge detection to extract comprehensive and accurate depth edges for patch deformation. In contrast, the previous work does not consider adopting monocular normal estimation for fine-grained depth edges fine tuning.
2) We improve cross-view depth reprojection to enhance the reliability of visibility maps and introduce the area-maximization strategy to address patch overlapping problems, while the previous work simply extract visibility maps without considering its robustness.
\yzl{3) We incorporate SHIQ for highlight correction, enabling propagation and refinement with highlight perception capacity, while the previous method ignores the impact of highlight regions on geometry consistency.}
4) We provide more experiments and ablation studies to exhibit the advancement and robustness of our approach.
In summary, our contributions are as follows:
\begin{itemize}
    \item We propose a depth-normal-edge prior that aligns coarse depth, normal and edge information through an erosion-dilation strategy, thereby generating fine-grained homogeneous boundaries to ensure stable deformation.
    \item \yzl{We implement cross-view depth reprojection and an area-maximization strategy to help reliably restore visible areas and effectively balance deformed patch.}
    \item \yzl{We obtain geometry consistency via normal aggregation and depth projection for both propagation and refinement, then employ SHIQ for highlight correction to enable the model with highlight perception capacity.}
    \item We achieve the state-of-the-art performance on ETH3D, Tanks \& Temples and Strecha datasets.
\end{itemize}

\section{Related Work}
\subsection{Traditional MVS Methods}
\subsubsection{PatchMatch MVS} 
PatchMatch \cite{PM} revolutionizes image matching by introducing a fast and efficient framework to find approximate correspondences between image patches through random initialization, propagation and refinement. 
PatchMatch Stereo \cite{PMS} extends the basic idea to multi-view stereo (MVS), offering a foundation for many subsequent developments. 
Gipuma \cite{Gipuma} enhances computational efficiency by implementing a red-black checkerboard propagation method, which allows for seamless GPU parallelism. 
ACMM \cite{ACMM} proposes a combination of adaptive sampling and multi-view consistency to improve depth reconstruction. 
Subsequent methods like ACMMP \cite{ACMMP} and HPM-MVS \cite{HPM-MVS}, refine this process by incorporating triangulation and KD-tree techniques for better handling of planar surfaces.
Furthermore, MG-MVS \cite{MG-MVS} and Pyramid \cite{Pyramid} introduce hierarchical processing structures, enabling the algorithm to better capture large-scale features and improve the receptive field for depth estimation. 
Despite these advancements, the insufficient receptive fields for patches continues to hinder the reconstruction of textureless areas.


\begin{figure*}
\centering
\includegraphics[width=\linewidth]{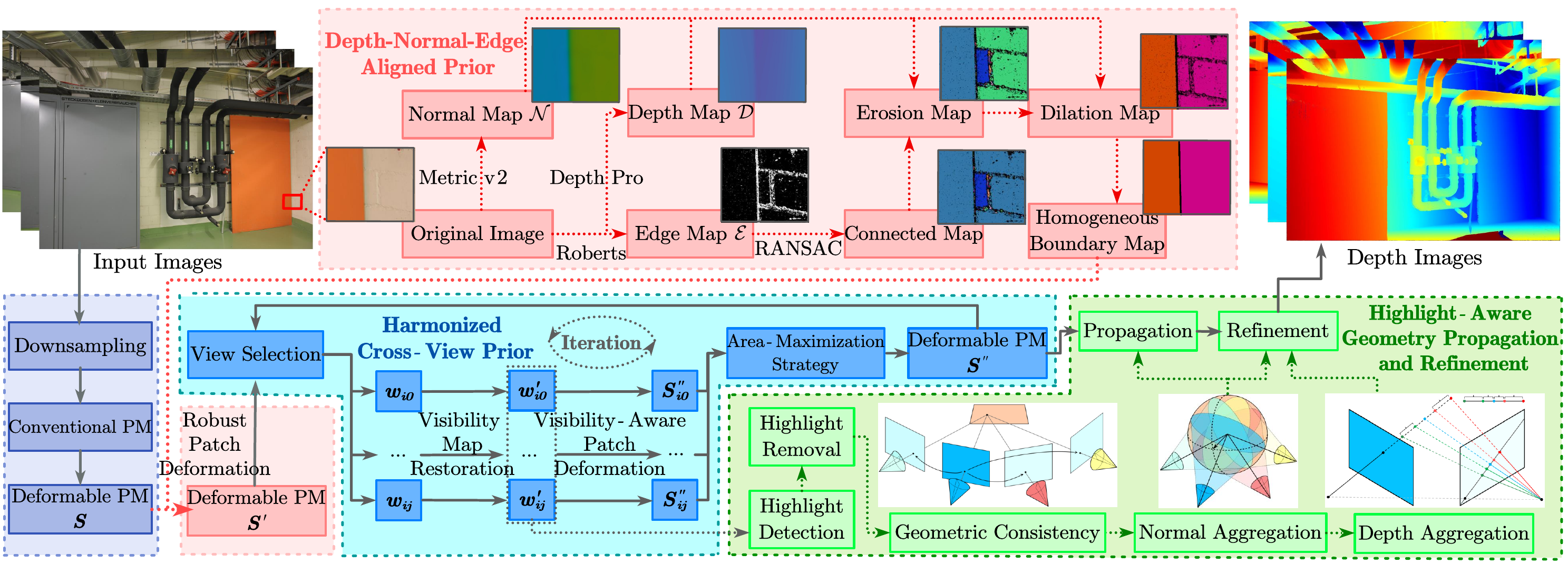}
\caption{An illustrated pipeline of DVP-MVS++. Specifically, we first introduce DepthPro, Metric3Dv2 and Roberts operator to respectively obtain depth, normal and edge maps, then adopt an erosion-dilation strategy to generate \textbf{depth-normal-edge aligned prior} for robust patch deformation. 
\yzl{Subsequently, we reform view selection for visibility map construction, then implement both depth reprojection and area-maximization strategy for visibility map restoration, which are then regarded as the \textbf{harmonized cross-view prior}.}
\yzl{Finally, we employ SHIQ for highlight correction, then respectively introduce visible normals aggregation and epipolar-based depth projection to acquire \textbf{highlight perecption capacity} for reconstruction.}
}
\label{fig:pipeline}
\end{figure*}

\subsubsection{Patch Deformation}
As an advancement within PatchMatch MVS, patch deformation enhances the patch receptive field by adapting the size and shape of patches for reconstruction. 
PHI-MVS \cite{PHI-MVS} utilizes dilated convolution to alter patch sizes based on local image context, while API-MVS \cite{API-MVS} applies entropy-based methods to adjust patch sampling intervals. 
However, large sampling intervals in both methods can lead to sparse sampling, reducing the effectiveness of depth matching.
SD-MVS \cite{SD-MVS} adopts instance segmentation to guide patch deformation, though its performance are limited by segmentation precision.
\yzl{In contrast, APD-MVS \cite{APD-MVS} splits unreliable pixel patches into multiple sub-patches to improve patch reliability, while the lack of depth edge constraints may cause issues like edge-skipping, especially in regions with abrupt depth changes.}

\subsection{Learning-based MVS Method}
MVSNet \cite{MVSNet} lays the foundation for learning-based MVS by leveraging differentiable 3D cost volumes to learn depth directly from 2D features.
Building on this, R-MVSNet \cite{R-MVSNet} and IterMVS-LS \cite{Iter-MVS} integrate GRU modules and lightweight probability estimators to improve regularization and depth estimation.
Moreover, both Cas-MVSNet \cite{Cas-MVSNet} propose efficient memory usage through a coarse-to-fine strategy, optimizing depth estimation across scales.
For feature aggregation, PatchMatchNet \cite{PatchMatchNet} combines PatchMatch with deep learning for acceleration during the reconstruction process.
MVSTER \cite{MVSTER} introduces an epipolar transformer to improve 2D-3D feature learning by leveraging cross-attention mechanisms, while EPP-MVSNet \cite{EPP-MVSNet} develops an epipolar-assembling module to assemble information from different resolutions.
Geo-MVSNet \cite{Geo-MVSNet} incorporates geometric priors through a two-branch fusion network, enhancing depth prediction accuracy.
RA-MVSNet \cite{RA-MVSNet} introduces point-to-surface distance integration within the cost volume to handle textureless areas effectively.
Despite these advances, challenges such as large-scale training datasets, excessive memory usage, and limited generalization remains significant obstacles for their large-scale applications, leading us to shift focus toward traditional MVS methods.

\section{Method}
\subsection{Overview}
Given a set of images $\mathcal{I} = \{I_i | i = 1, ..., N \}$ and their corresponding camera parameters $\mathcal{P}=\{K_i, R_i, T_i \mid i=1 \cdots N\}$, we sequentially select a reference image $I_{\mathrm{ref}}$ from $\mathcal{I}$ and reconstruct its depth map by performing pairwise matching with the remaining source images $I_{\mathrm{src}}(\mathcal{I} - I_{\mathrm{ref}})$. The overall pipeline of our method is shown in Fig. \ref{fig:pipeline}, and the details of each component will be discussed in the following sections.

\subsection{Preliminary Area}
The traditional PM method projects a fixed-size patch from the reference image onto the corresponding patch in the source image based on a given plane hypothesis. It then calculates the similarity between the two patches using the NCC matrix~\cite{COLMAP}. More specifically, for a reference image $I_i$ and a source image $I_j$, for each pixel $p$ in $I_i$, we first generate a random plane hypothesis $(\mathbf{n}^T, d)$, where $\mathbf{n}$ is the normal vector, and $d$ is the depth. Using homography mapping \cite{Accurate}, we can compute the projection matrix $H_{ij}$ for the plane hypothesis $(\mathbf{n}^T, d)$ at pixel $p$ between images $I_i$ and $I_j$. Then, $H_{ij}$ is used to project the fixed-size patch $B_p$ centered at pixel $p$ in the reference image $I_i$ onto the corresponding patch $B_p^j$ in the source image $I_j$. The matching cost is calculated as the NCC score between $B_p$ and $B_p^j$, which is defined as:
\begin{equation}
m_{ij}\left(p, B_p\right)=1-\frac{cov\left(B_p, B_p^j\right)}{\sqrt{cov\left(B_p, B_p\right) cov\left(B_p^j, B_p^j\right)}},
\end{equation}
where $cov$ denotes the weighted covariance \cite{COLMAP}. The multi-view aggregated cost is then defined as:
\begin{equation}
m\left(p, B_p\right)=\frac{\sum^{N-1}_{j=1} w_{ij}(p) \cdot m_{ij}\left(p, B_p\right)}{\sum^{N-1}_{j=1} w_{ij}(p)},
\end{equation}
where the weight $w_{ij}(p)$ is derived from the view selection strategy~\cite{ACMMP}. Finally, by propagating and refining the hypotheses for each pixel, we select the hypothesis with the lowest matching cost as the final result.

Alternatively, the deformable PM method \cite{APD-MVS} splits the patch of an unreliable pixel into multiple reliable sub-patches, with the center of each sub-patch regarded as an anchor pixel. For each unreliable pixel $p$, its cost in deformable PM is:
\begin{equation}
m_{ij}(p, S) = \lambda m_{ij}(p, B_p) + (1 - \lambda) \frac{\sum_{s \in S} m_{ij}(s, B_s)}{|S|},
\label{PM}
\end{equation}
where $S$ represents the set of all anchors, and $B_s$ is the sub-patch centered around each anchor pixel $s \in S$. In practice, the patch sizes for both $B_p$ and $B_s$ are $11 \times 11$, with sampling intervals of 5 and 2, respectively. $\lambda = 0.25$, $|S| = 8$.





\subsection{Depth-Normal-Edge Aligned Prior}
\begin{figure*}
\centering
\includegraphics[width=\linewidth]{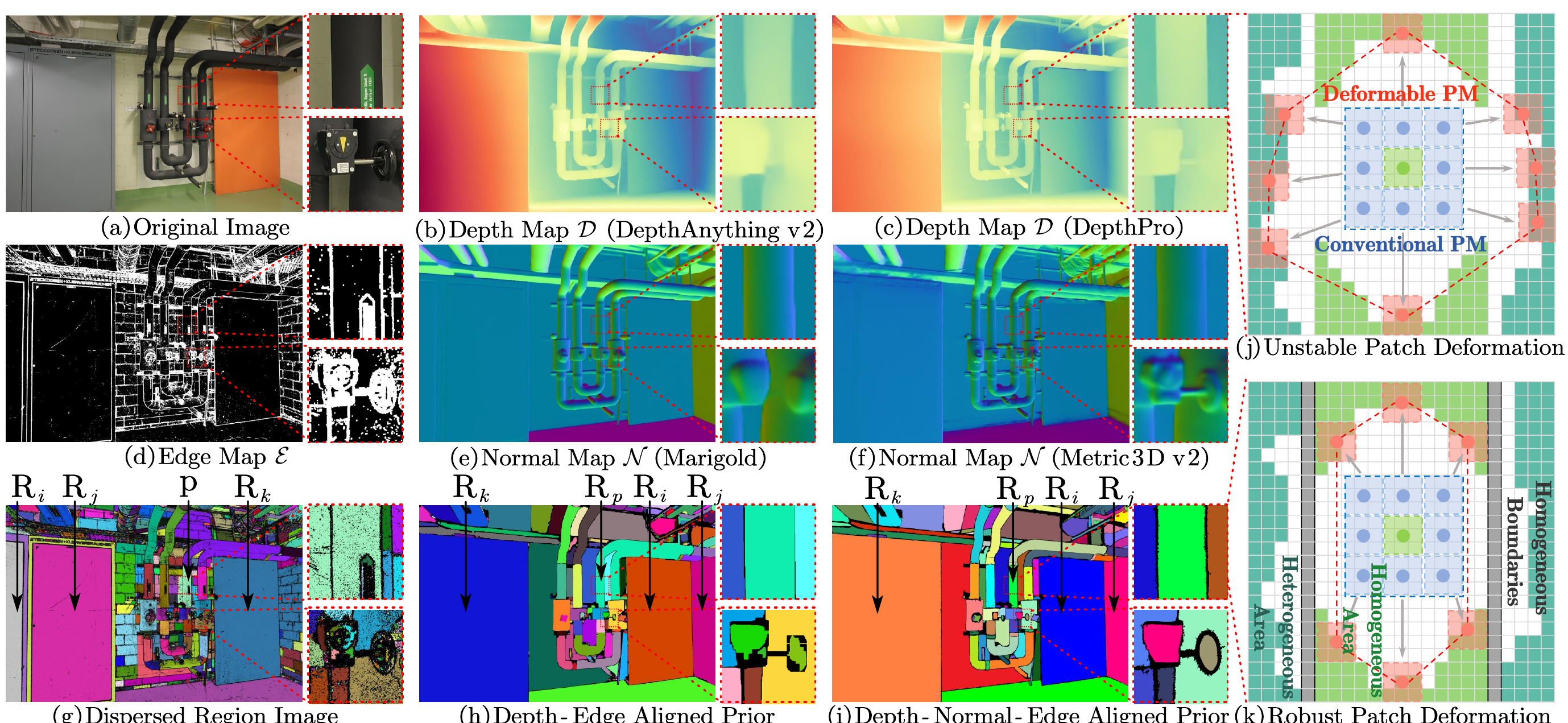}
\caption{
Depth-Normal-Edge Aligned Prior. In (b) and (c), areas with lower color temperature show smaller depths, while higher color temperatures indicate greater depths. In (e) and (f), different colors correspond to distinct normal planes.
In (c), Edges are highlighted in white, with black constituting dispersed regions in (g).
Moreover, different colors represent different dispersed regions in (g) and homogeneous areas in (h) and (i), with black denoting homogeneous boundaries. 
In (j) and (k), green, blue and red block respectively represent the central pixel, corresponding neighbors in conventional PM and anchors in deformable PM. 
Moreover, cyan, green and gray backgrounds respectively indicate heterogeneous areas, homogeneous areas and homogeneous boundaries.
}
\label{fig: DEP}
\end{figure*}

Recent methods have adopted patch deformation to expand the receptive field of patches, thereby effectively reconstructing unreliable pixels in textureless areas. However, they overlook the crucial depth edge constraint required for deformation stability. As illustrated in Fig.~\ref{fig: DEP} (j), the lack of depth edge constraint leads to edge-skipping in the deformed patch, causing it to cover heterogeneous regions with depth discontinuities and potentially resulting in estimation errors.

To address this challenge, DVP-MVS~\cite{DVP-MVS} combines monocular depth estimation and edge detection to extract depth edges for patch deformation. However, relying solely on monocular depth for depth edge extraction proves to be unreliable.
When faced with outdoor scenes with a wide depth range, a single threshold cannot adequately assess depth continuity, making it challenging to extract comprehensive and reliable depth edges.
For instance, a small threshold is suitable for identifying depth continuity of nearby objects, but a larger threshold is instead necessary for distant ones.
While monocular normal estimation can effectively solve this issue.
By capturing normal differences between adjacent pixels, it can accurately identify depth edges regardless of current depth value, making it accommodate to various scenarios. Furthermore, compared to depth differences, normal differences provide a finer granularity for edge alignment, enabling more robust patch deformation.

Therefore, to fully exploit depth edges, we combine both monocular depth estimation and monocular normal estimation to generate a depth-normal-edge aligned prior for robust patch deformation.
Specifically,  given the original image $I$ in Fig.\ref{fig: DEP} (a), we initially apply DepthPro~\cite{DepthPro} to initialize the depth maps $\mathcal{D}$ shown in Fig. \ref{fig: DEP} (c).
Unlike DVP-MVS which adopts DepthAnythingV2~\cite{depany2} to generate depth maps in Fig. \ref{fig: DEP} (b), DepthPro provides more reliable depth maps with finer details, including hair and intricate structures, as demonstrated in Fig. \ref{fig: DEP} (c), making it better suited for depth edges extraction within complicated scenes.
Moreover, DepthPro performs faster than DepthAnythingV2, making it ideal for real-time applications. A detailed comparison is provided in its paper~\cite{DepthPro}.

Moreover, we further employ Metric3Dv2 on the image $I$ to initialize the normal maps $\mathcal{N}$ shown in Fig.~\ref{fig: DEP} (f). 
Unlike other monocular normal estimation algorithms like Marigold~\cite{Marigold} and ND-Depth~\cite{NDDepth}, Metric3Dv2 can effectively generate the most realistic surface normals and exhibits excellent generalization across diverse scenarios.
By comparing Fig. \ref{fig: DEP} (f) with Fig. \ref{fig: DEP} (e), we observe that the normal maps $\mathcal{N}$ generated by Metric3Dv2 can effectively preserve the geometric structure with finer details, which can provide a more reliable guidance for subsequent depth edge extraction and alignment.

Meanwhile, we apply the Roberts operator on the image $I$ to extract coarse edge maps $\mathcal{E}$ in Fig. \ref{fig: DEP} (d) and all dispersed regions $\mathcal{R}$ visualized in different colors in Fig. \ref{fig: DEP} (g). This process can be described as: $I=\mathcal{E} \cup \mathcal{R}$. 
For convenience, we define the heterogeneous region $H=\{p \in \Omega \mid \nabla d_p > \epsilon \}$, the homogeneous region as $M=\{p \in \Omega \mid \nabla d_p \leq \epsilon \}$ and the homogeneous boundary as $E_{i,j}=M_i \cap M_j$, where $\nabla d_p$ represents the depth gradient of pixel $p$, and $\epsilon$ is set to 0.005.
Therefore, to facilitate the alignment of depth maps, normal maps and edge maps for generating fine-grained homogeneous boundaries, we propose a region-level erosion-dilation strategy to distinguish and aggregate homogeneous regions.

\yzl{Specifically, for each dispersed region $R_k$ in Fig. \ref{fig: DEP} (g)}, if its size exceeds $\eta$, we combine its depth and region information in $\left[\mathcal{D}, \mathcal{R}\right]$ to perform region-wise planarization using RANSAC, resulting in the estimated plane $\pi_k = (n_k, d_k)$. 
We also define the inlier ratio $r_k$ as the proportion of inlier pixels to the total number of pixels during each RANSAC planarization. Consequently, we obtain the set $\mathcal{R} = \{R_k, \pi_k, r_k | k = 1, ..., N \}$.
During the erosion stage, we apply intra-regional erosion to separate regions belonging to heterogeneous areas. For example, region $R_k$ is pre-divided into two sub-regions $R_i$ and $R_j$ through erosion. We then reapply planarization on each sub-region to respectively obtain their respective estimated planes $\pi_i = (n_i, d_i)$, $\pi_j = (n_j, d_j)$, and inlier ratios $r_i$, $r_j$. 
Then we define the plane similarity function $\varPsi(\pi_i, \pi_j)$ as:
\begin{equation}
\varPsi(\pi_i, \pi_j) = \left(n_i \cdot n_j\right) - \min(1, |d_i - d_j|),
\end{equation}
where when both normals of planes $\pi_i$ and $\pi_j$ are dissimilar and their depth difference is large, the dot product $n_i \cdot n_j$ will be small, leading to a low value for $\varPsi(\pi_i, \pi_j)$, and vice versa.

Apart from this, we can derive all boundary pixels between the sub-regions $R_i$ and $R_j$ and merge them into a boundary pixel set, denoted as $\mathcal{P}$. We then define the normal similarity function $\varPhi(n_p)$ for each boundary pixel $p \in \mathcal{P}$ as:
\begin{equation}
\varPhi(n_p) = \min _{q \in N(p)}\left(n_p \cdot n_q\right),
\end{equation}
where $N(p)$ denotes the 4-neighborhood pixels of $p$. Here, the direction of the largest normal change for pixel $p$ corresponds to the minimum dot product with the normals of its neighboring pixels. Therefore, when the normal of pixel $p$ undergoes significant change, $\varPhi(n_p)$ will be small, and vice versa.
Note that since there might be slight deviations between the detected boundaries $\mathcal{P}$ and the actual depth edges, for each boundary pixel $p \in \mathcal{P}$,  we extend its search area within a specific range around its four neighboring directions when computing $\varPhi(n_p)$. Such operation effectively enlarge pixel-wise receptive field, ensuring the accurate identification of depth boundaries.
Finally, we consider erosion is effective and perform region division $R_k \xrightarrow{}\left\{R_i, R_j\right\}$ when the following condition holds:
\begin{equation}
\varPsi(\pi_i, \pi_j) \leq \varphi \text { s.t.} \frac{r_i+r_j}{2 r_k} \geq \gamma \text { and }\frac{1}{|\mathcal{P}|}\sum_{p \in \mathcal{P}} \varPhi(n_p) \leq \phi.
\label{erosion}
\end{equation}
Respectively, when the plane similarity function $\varPsi(\pi_i, \pi_j)$ is less than the threshold $\varphi$ indicates that both sub-regions are dissimilar, suggesting that they belong to heterogeneous areas. 
If the sum of two inlier ratios $r_i+r_j$ exceeds twice the original inlier ratio $r_k$, it means a superior planarization after erosion. 
Moreover, once the average normal similarity function $\varPhi(n_p)$ is lower than the threshold $\phi$, it indicates noticeable normal fluctuations along the boundary between the two sub-regions.

On the other hand, in the dilation stage, we perform inter-regional dilation to merge regions that belong to homogeneous regions. For instance, regions \yzl{$R_i$} and \yzl{$R_j$} are pre-merged into an aggregated region \yzl{$R_k$} through dilation. We then determine dilation is valid and perform region merging \yzl{$\{R_i, R_j\} \xrightarrow{\text{}} R_k$} under the following condition:
\begin{equation}
\varPsi(\pi_u, \pi_v) \geq \varphi \text { s.t. } r_u, r_v \geq \kappa \text { and }\frac{1}{|\mathcal{P}|}\sum_{p \in \mathcal{P}} \varPhi(n_p) \geq \phi,
\label{erosion} 
\end{equation}
Similarly, if two estimated planes $\pi_u$ and $\pi_v$ are both reliable and similar, with the normals of their boundaries exhibit nearly no variations, we consider that regions \yzl{$R_i$} and \yzl{$R_j$} should be merged together as homogeneous regions \yzl{$R_k$}.

Following erosion-dilation strategy, we further apply pixel-wise filtering to refine fine-grained homogeneous boundaries. 
Specifically, when a pixel $p=(x, y, d_p)$ with depth $d_p$ is adjacent to region \yzl{$R_p$} with estimated plane $\pi_k = (\mathbf{n}_k, d_k)$, we consider $p$ to belong to \yzl{$R_p$} if the following condition is met:
\begin{equation}
\frac{| n_k \cdot p + d_k |}{|n_k|} \leq \delta \text{ s.t. } r_k \geq \kappa \text { and } \varPhi(n_p) \geq \phi,
\label{filtering}
\end{equation}
in Eq. \ref{filtering}, if the 3D position of $p$ is close to the estimated plane and there is no significant change in its normal, then $p$ should be considered part of region \yzl{$R_p$}. Through the above-mentioned strategy, we subtly combine the reliability of monocular depth in constructing plane regions and the accuracy of monocular normals for edge detail verification, thus extracting reliable depth edges for subsequent patch deformation. Compared to the depth-edge aligned prior produced by DVP-MVS \cite{DVP-MVS} shown in Fig. \ref{fig: DEP} (h), our depth-normal-edge aligned prior shown in Fig. \ref{fig: DEP} (i) can perceive fine-grained homogeneous boundaries and better distinguish different homogeneous regions, making it more suitable for guiding patch deformation.

After obtaining the depth-normal-edge aligned prior, we adopt it as guidance to limit patch deformation within homogeneous regions. Specifically, as depicted in Fig.~\ref{fig: DEP} (k), for each unreliable pixel $p$, we first adopt patch deformation to acquire its anchor collection $S$. Then for each anchor $s_i \in S$, we retain it only if it belongs to the same region as pixel $p$; otherwise, we discard it. This operation can be expressed as:
\yzl{
\begin{equation}
S^{\prime} = \{ s_i \in S \mid s_i \in R_k \},
\label{depth-normal-edge}
\end{equation}
}
where $S^{\prime}$ denotes the updated anchor collection and \yzl{$R_k$} is the homogeneous regions that $p$ belongs to. Finally, the new collection $S^{\prime}$ will replaces the old one $S$ for further patch deformation, ensuring that deformed patches stay within homogeneous regions and preventing unexpected edge-skipping. 




\subsection{Harmonized Cross-View Prior}
In addition to the edge-skipping problem mentioned earlier, another significant challenge during patch deformation is the visibility discrepancy and occlusion caused viewpoint change. As illustrated in Fig.~\ref{fig: view}, certain areas of the reference image in (a) correspond to invisible areas of the source image in (b). However, the deformed patch in (e) erroneously includes these invisible areas during  matching cost, thus leading in potential distortions. To tackle this, we aim to progressively incorporate cross-view prior to achieve visibility-aware patch deformation.

Moreover, as shown in Fig.~\ref{fig: view} (e), since anchor selection exists somewhat uncertainty, patch deformation often results in red pixels, i.e. anchors, being selected too close to each other, causing receptive field overlapping during cost computation. Such phenomenon not only leads to an insufficient receptive field but also causes an imbalanced weights for the receptive field, both of which contribute to potential matching ambiguity. To address this, we propose an area-maximization anchor selection strategy for optimization, thus acquiring a harmonized cross-view prior that guarantees the maximization of effective coverage of patch reception field within homogeneous regions.

\subsubsection{Visibility Map Restoration}
\begin{figure}
\centering
\includegraphics[width=\linewidth]{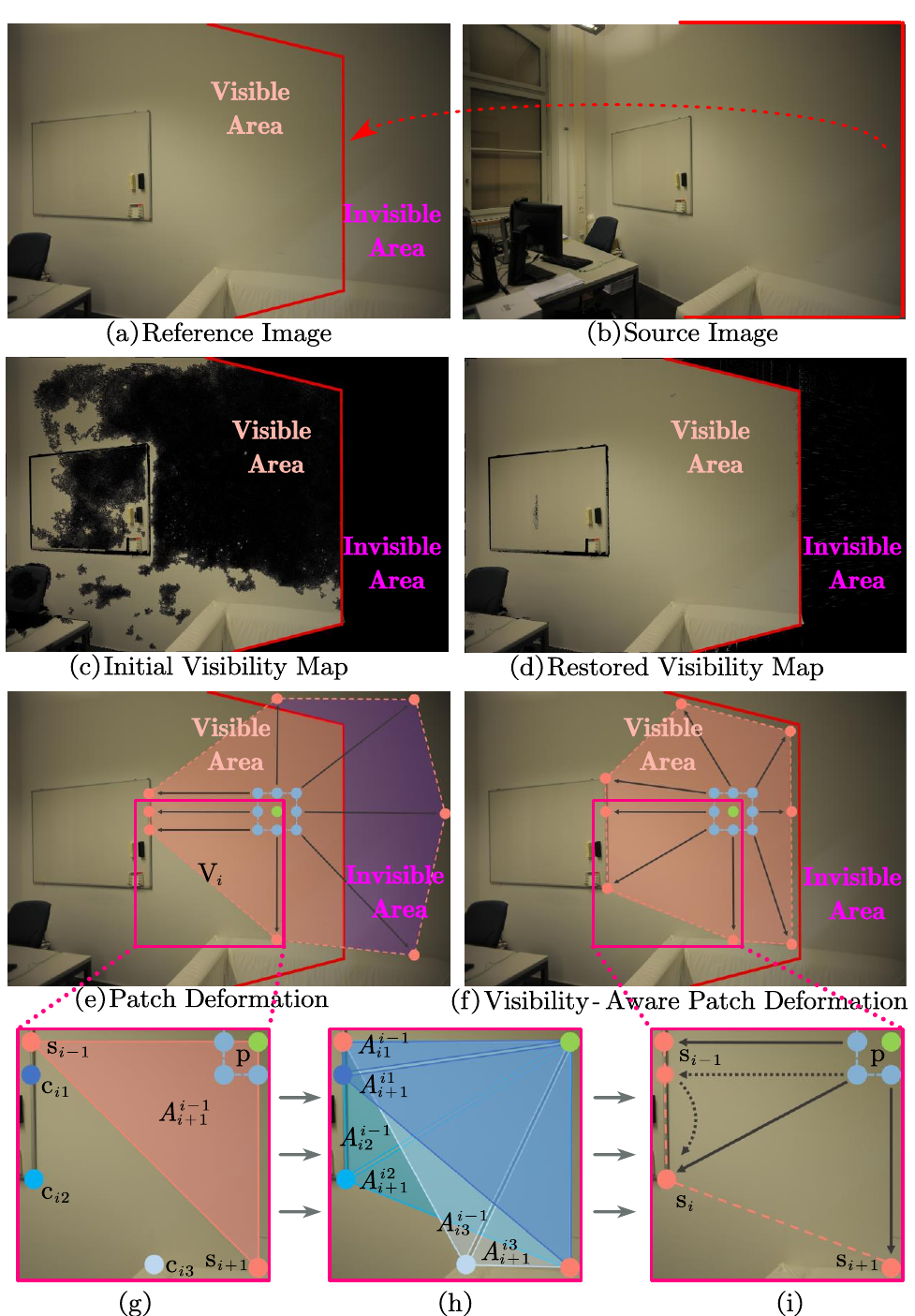}
\caption{
\yzl{Cross-View Prior. The red line in (a) separates visible and invisible areas of (a) within (b). In (c) and (d), black indicates pixels judged invisible by the view selection strategy. In (e) and (f), green, blue, and red indicate the central pixel, conventional PM neighbors, and deformable PM anchors, respectively. Subfigures (g), (h), and (i) illustrate the Area-Maximization Strategy, focusing on a localized example where the anchor $S_{i}$ is adjusted to maximize coverage while 
$S_{i-1}$ and $S_{i+1}$ remain fixed. Subfigure (f) shows the global result of this strategy applied to all anchors, whereas subfigure (i) zooms into the localized optimization process. This distinction ensures clarity between the macroscopic outcome and the microscopic implementation.
}
}
\label{fig: view}
\vspace{-1.0em}
\end{figure}

For each pixel $p$ in the reference image $I_i$, we begin by initializing its visibility weights $w_{ij}(p)$ for the corresponding source images $I_j$ using the view selection strategy from~\cite{ACMMP}. 
Here the weights $w_{ij}(p)$ are not only used in cost aggregation but are also reformulated into visibility maps that provide patch deformation with cross-view visibility perception. 
According to~\cite{ACMMP}, $w_{ij}(p)$ is computed by comparing costs against fixed thresholds.
As a result, pixels with low costs are more likely to be considered visible, while those with high costs are often deemed invisible.
Thereby, as shwon in Fig. \ref{fig: view} (c), in textureless areas characterized by high costs, regions that were originally visible may be incorrectly judged as invisible.
Consequently, relying solely on cost for visibility restoration is unreliable, as it fails to properly distinguish between well-textured and textureless areas.

Visibility essentially refers to the existence of corresponding pixel pairs between images, meaning one pixel can be mapped to another via projection. Therefore, we attempt to employ the cross-view depth reprojection $e(p)$~\cite{ACMMP} as the post-verification for visibility map restoration. 
Through back-and-forth projection, $e(p)$ can not only verify the visibility between pixel pairs but also assess the robustness of depth estimation.

Specifically, for each pixel $p_i$ in the reference image $I_i$, we first use its depth $d_i$ to project it into pixel $p_j$ in the source image $I_j$. We then reproject $p_j$ from $I_j$ back to $I_i$ using its corresponding depth $d_j$, which yields the reprojection pixel $p^{\prime}_i$. 
We then define $e(p)=\left\|p^{\prime}_i-p_i\right\|$ and consider $p_i$ in $I_i$ is also visible in $I_j$ when $e(p) \leq \varepsilon$.
This process generates both the restored weight $w_{ij}^{\prime}(p)$ and the visibility map shown in Fig.~\ref{fig: view} (d), which effectively restores visibility in textureless areas.

However, since the calculation of $e(p)$ is highly dependent on the depth reliability of individual pixel, their results tend to be error-prone. To tackle this limitation,  we make some improvements.
Specifically, during each projection, the algorithm will construct a $11 \times 11$ window centered around the projected pixel. Then the depth of pixel with the lowest cost within the window will substitute the depth of projected pixel for calculation, thereby enhanceing the reprojection reliability.

\yzl{
Note that the depth adopted for computing reprojection errors are generated through an iterative refinement similar to ACMMP~\cite{ACMMP}, where the initial stage computes reprojection errors based on visibility maps derived from view selection weights without requiring precomputed depth. Subsequent iterations progressively refine depths using results from earlier stages, ensuring consistency across views and eliminating reliance on external depth sources for computations. 
}

\subsubsection{Visibility-Aware Patch Deformation}
After obtaining the visibility map, we treat it as the cross-view prior to iteratively updating patch deformation.
Specifically, for each unreliable pixel $p$ in the reference image $I_i$, we first adopt patch deformation to generate the initial anchor collection $S$, which is then guided by the depth-normal-edge aligned prior to produce the new collection $S^{\prime}$ via Eq. \ref{depth-normal-edge}. Subsequently, for each anchor $s_i \in S^{\prime}$, we retain it if it is visible in the source image $I_j$; otherwise, it is discarded. Combining with Eq. \ref{depth-normal-edge}, the view-level anchor collection is now formulated by:
\yzl{
\begin{equation}
S_{ij}^{\prime\prime} = \{ s_i \in S^{\prime} \mid s_i \in R_k, w_{ij}(p) > 0 \}.
\end{equation}
}
Then $S_{ij}^{\prime}$ will replace $S$ in Eq. \ref{PM} to calculate the matching cost $m_{ij}(p, S_{ij}^{\prime})$, which in turn, can further optimize $w_{ij}^{\prime}(p)$ during the view selection process, thus iteratively promoting patch deformation in Fig.~\ref{fig: view} (f) with visibility perception. The final multi-view aggregated cost  $m(p,S^{\prime\prime})$ is then defined by:
\begin{equation}
m\left(p, S^{\prime\prime}\right)=\frac{\sum^{N-1}_{j=1} w_{ij}^{\prime}(p) \cdot m_{ij}\left(p, S_{ij}^{\prime\prime}\right)}{\sum^{N-1}_{j=1} w_{ij}^{\prime}(p)}.
\end{equation}
Moreover, we further apply the new deformed patch composed by $S_{ij}^{\prime\prime}$ during the propagation process. Each $s_i \in S_{ij}^{\prime\prime}$ will be adopted as candidate plane hypotheses to be propagated to the unreliable pixel, thereby accelerating the convergence.

\begin{figure*}
\centering
\includegraphics[width=\linewidth]{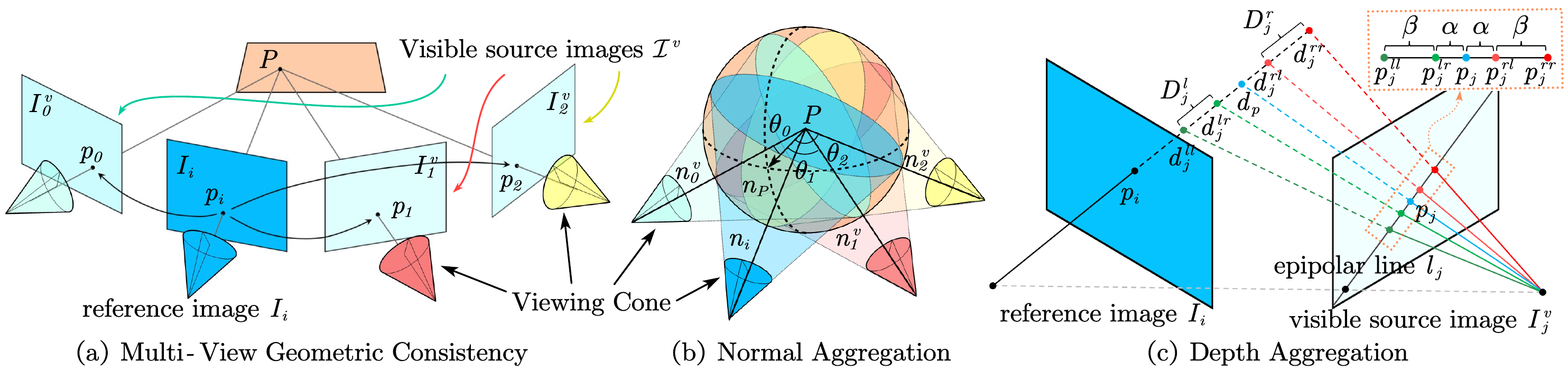}
\caption{\yzl{Geometry-Driven Propagation and Refinement. In (a) and (b), the blue view cone corresponds to the reference image $I_i$, while green, red, and yellow view cones correspond to visible source images $I^v_0$, $I^v_1$ and $I^v_2$. In (c), by projecting the extended line from the camera center to pixel $p_i$ on the reference image $I_i$, we can obtain the epipolar line $l_j$ on the visible source image $I^v_j$.}
}
\label{fig: epip}
\end{figure*}

\subsubsection{Area-Maximization Strategy}
As shown in Fig.~\ref{fig: view} (e), during patch deformation, the algorithm first divides its surrounding area into $|S|$ sectors, then selects X candidate within each sectors and ultimately chooses them as anchor pixels for reconstruction. 
However, in practice, anchor pixels are often selected too close between each other, leading to an patch overlapping phenomenon. Thereby, it not only makes receptive fields insufficient but also causes an imbalance weights between them, leading to potential matching ambiguities. 

To address this, we propose an area-maximization strategy for anchor selection. As shown in Fig.~\ref{fig: view} (g), given unreliable pixel $p$, we first consider all its reliable candidates $\mathcal{C}_i = \{c_{ij} \mid j = 1, ..., N\}$ in the $i^{th}$ sector $V_i$ during patch deformation. Then, we respectively include each candidate $c_{ij} \in \mathcal{C}_i$ into the anchor collection $S$ and calculate the growth area of the patch, termed as $\Delta A_{ij}$. Then the included candidate with the maximum $\Delta A_{ij}$ is selected as the ultimate anchor pixel $s_i$:
\begin{equation}
s_i = \mathop{\text{argmax}}\limits_{c_{ij} \in \mathcal{C}_i} \Delta A_{ij}
\label{max}
\end{equation}
Specifically, given a candidate $c_{ij}$ with its neighboring anchor pixels $s_{i-1}$, $s_{i+1}$, we first calculate the area of the triangle formed by $s_{i-1}$, $s_{i+1}$, $p$, denoted as $A^{i-1}_{i+1}$. Subsequently, after including candidate $c_{ij}$ into the anchor collection $S$, we can compute the areas of two sub-triangles formed by $s_{i-1}$, $c_{ij}$, $p$, and $s_{i+1}$, $c_{ij}$, $p$, denoted as $A^{i-1}_{ij}$ and $A^{ij}_{i+1}$, respectively. The growth area is thus derived as $\Delta A_{ij} = A^{i-1}_{ij} + A^{ij}_{i+1} - A^{i-1}_{i+1}$. By solving Eq. \ref{max}, we ensure the selected anchor pixel maximizes the local coverage, promoting a balanced distribution of the patch receptive field within homogeneous region. 
\yzl{Therefore, the proposed area-maximization strategy realizes a harmonized cross-view prior, equipping patch deformation with balanced and comprehensive perceptual capabilities.}








\yzl{\subsection{Geometry Propagation and Refinement}}
Previous algorithms \cite{ACMMP} select optimal neighbor hypotheses for propagation and randomly generate new hypotheses through perturbation for refinement.
However, since each pixel is visible in different viewpoints, each viewpoint can only view normal above $90^{\circ}$ from its normal direction. Therefore, we can leverage such geometric constraint to limit pixel-wise hypothesis range.
Specifically, we attempt to regard the restored weight $w_{ij}^{\prime}(p)$ as the pixel-wise visibility prior to adaptively constrain pixel-wise hypothesis range for propagation and refinement, thereby improving hypothesis accuracy during reconstruction. 

Moreover, since the PM algorithm estimates depth based on RGB image colors, it faces matching failures when encountering highlight regions, whose colors change with the viewpoint, and may further affecting the reliability of visibility prior.
\yzl{To resolve this, we use SHIQ~\cite{ACMMP}, a highlight correction algorithm, to adjust over-saturated or reflective regions in images. This improves depth estimation by ensuring geometric consistency even in challenging lighting conditions.
}

\subsubsection{Normal Aggregation}
We first utilize the visibility maps to aggregate multi-view normals, thereby restricting the normals to their respective visible ranges.
As illustrated in Fig. \ref{fig: epip} (a), for each pixel $p_i$ in the reference image $I_i$, we first adopt its restored visibility maps to identify all visible source images $\mathcal{I}^v = \{I^v_j | j = 1, ..., N - 1, w_{ij}^{\prime}(p) > 0\}$. Then through depth projection, we can obtain the 3D point $P$ and mapping pixels $p_j$ corresponding to each visible source image $I^v_j$. Since each visible source image $I^v_j$ in $\mathcal{I}^v$ possesses a visible hemisphere in its own viewing cone, and $P$ is visible to all visible source images $\mathcal{I}^v$, we can aggregate all the visible hemispheres in $\mathcal{I}^v$ to constrain 3D point's normal range.

Specifically, we first obtain the normal direction of viewing cone  $n^v_j$ for each visible source image $I^v_j$ by connecting its camera center with the corresponding mapped pixels $p_j$. Subsequently, as shown in Fig. \ref{fig: epip} (b), for each visible source image $I^v_j$, we ensure the angle $\theta_j$ between its normal direction $n^v_j$ and the normal $n_P$ of 3D point $P$ must be greater than $90^{\circ}$ (i.e., $n_P \cdot n^v_j \leq 0$). 
Ultimately, by reprojecting the aggregated normal range of 3D point $P$ back onto pixel $p_i$ in the reference image $I_i$, we enforce a constraint on its normal range $n_i$. 
This aggregated normal range is then adopted as the constraint during the propagation and refinement steps to help enhancing the robustness of normal hypotheses.

\begin{figure*}
\centering
\includegraphics[width=\linewidth]{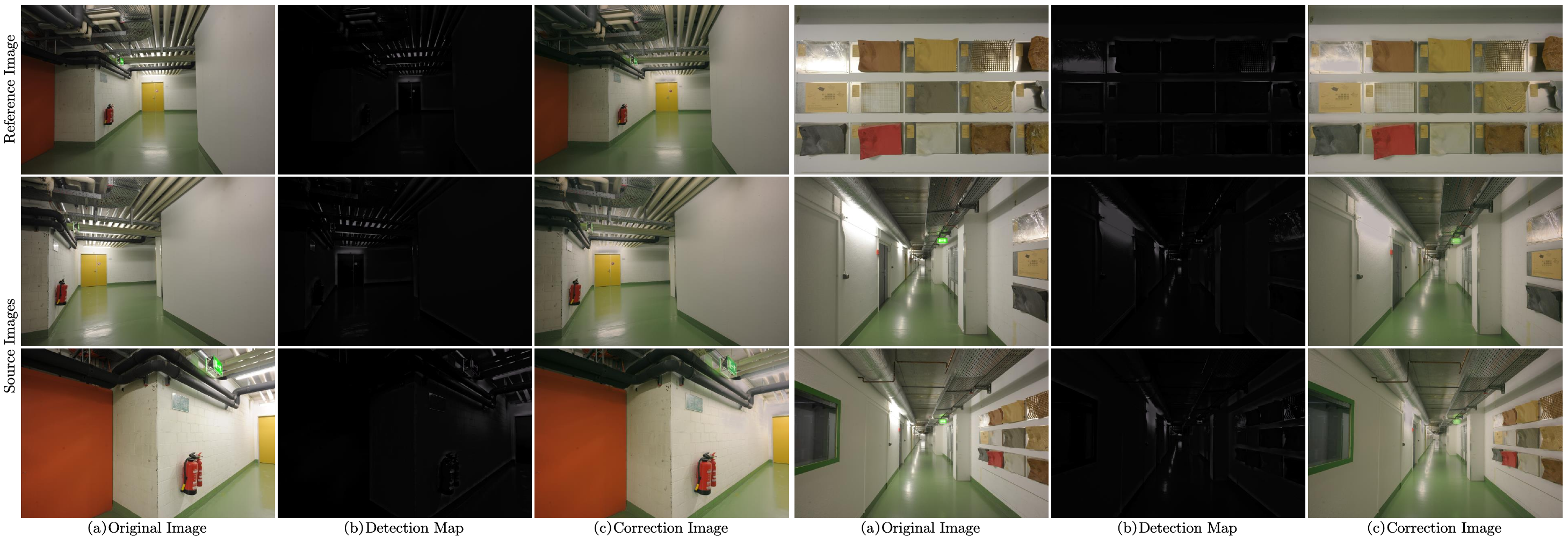}
\caption{Highlight correction. By removing the detected highlight area (b) of the original image (a), we acquire the corresponding correction image (c).
}
\label{fig: specular}
\end{figure*}

\subsubsection{Depth Aggregation}
Conventional fixed depth intervals encounter two primary issues during the refinement stage: 
$1)$ when depth intervals are evenly spaced in the reference image $I_i$ and projected onto the source image $I_j$, the spacing between the corresponding points along the epipolar line becomes irregular, progressively narrowing as depth increases;
$2)$ for each depth interval along the epipolar lines of the reference image $I_i$, their projected intervals along the epipolar is variable across different source images.

To resolve these issues, we employ inverse projection and aggregate fixed-length samples along the epipolar lines from multiple views to adjust the depth intervals.
As shown in Fig. \ref{fig: epip} (c), for each pixel $p_i$ in the reference image $I_i$, we first locate its corresponding pixel $p_j$ along the epipolar line $l_j$ in the visible source image $I^v_j$, as described earlier. 
Subsequently, four pixels are selected at distances $[\alpha, \alpha+\beta]$ from $p$ on both sides along epipolar line $l_j$, which define the extremes of minimum and maximum disturbance range of the mapping pixels. These four extremal pixels are then inversely projected into $I_i$ to accordingly obtain four depth values, thereby forming two intervals for refinement: $D^l_{j} = (d^{ll}{j}, d^{lr}{j})$ and $D^r_{j} = (d^{rl}{j}, d^{rr}{j})$. 
This adaptive approach guarantees that the refined mapping pixels can be displaced appropriately along the epipolar line, free from the constraints of the current depth.

Additionally, to assure that mapping pixels undergo significant displacements across as many source images as possible, we further aggregate the depth intervals from all visible source images $\mathcal{I}^v$. 
To achieve this, we calculate the boundaries of the aggregated intervals by selecting the $\mu^{th}$ smallest value from all of maximum disturbance extremes (i.e., $d^{ll}_{j}$ and $d^{rr}_{j}$), and the $\mu^{th}$ largest value from all minimum disturbance extremes (i.e., $d^{lr}_{j}$ and $d^{rl}_{j}$), making the aggregated interval defined by: 
\begin{equation}
\left(\min_{1\leq j\leq \mu} d^{ll}_{j}, \max_{1\leq j\leq \mu} d^{lr}_{j}\right), 
\left(\max_{1\leq j\leq \mu} d^{rl}_{j}, \min _{1\leq j\leq \mu} d^{rr}_{j}\right).
\end{equation}
Further clarification regarding this formula can be found in the supplementary material.
Ultimately, the aggregated intervals will replace the fixed ones during the local perturbation of refinement, thus increasing the reliability of depth hypotheses.

\yzl{\subsubsection{Highlight Perception Capacity}}
As shown in Fig. \ref{fig: specular}, since the highlight regions exhibit strong reflections, its original colors are replace with intense white. This prevents the algorithm from making accurate depth estimations, leading to potential reconstruction distortions.
To address this, we introduce SHIQ for highlight correction, which can effectively leverage reliable information from multi-view images for both highlight regions detection and correction. 

Specifically, as shown in Fig. \ref{fig: specular}, given the reference image and the source image, we first adopt SHIQ to respectively obtain their detection maps and correction images.
Since correction images can can effectively restore the original highlight regions, we then adopt them to replace all original images for patch matching, thereby mitigating the impact of highlight regions on reconstruction failure. 

However, correction images may not always be  reliable for patch matching, especially when the highlight region is well-textured. In such cases, SHIQ struggles to restore the complex color details of these regions. 
To overcome this, we subtly adjust our reconstruction strategy for highlight regions, making it more flexible for different situations. 

Specifically, for each pixel $p$ within the highlight region of the reference image $I_i$,  if it is reliable, we do not reconstruct it for safety, as it can still be reconstructed from other views. Differently, if it is unreliable, we remove its central receptive field and retain only the surrounding ones for patch matching. This modification alters deformable PM from equation (3) to:
\begin{equation}
m_{ij}(p, S) =\cancel{ \lambda m_{ij}(p, B_p)} + \cancel{(1 - \lambda)} \frac{\sum_{s \in S} m_{ij}(s, B_s)}{|S|}.
\end{equation}
Moreover, for each pixel $p$ within the highlight region of the source image $I_j$, if it is deemed visible through our visibility map restoration and is reliable, we consider it as invisible to avoid potential erroneous depth estimation. Conversely, if it is unreliable, then it won't be adopted for patch deformation, so we don't need to make any changes.
With these adjustments, we effectively solve the issues caused by highlight regions in both patch matching and visibility priors, providing effective guidance for both normal and depth aggregations.








\section{Experiments}
We test our approach on three distinct datasets: the ETH3D high-resolution benchmark \cite{ETH3D}, the Tanks \& Temples (TNT) benchmark \cite{TNT} and the Strecha dataset \cite{strecha}. Our evaluation begins with both qualitative and quantitative comparisons of our point clouds against other methods across these datasets. Following this, we perform ablation studies on each proposed module and key parameters to validate their effectiveness and justify our choices. Additionally, we assess the efficiency of our method by comparing GPU memory usage and runtime. The extensive results demonstrate that our method achieves state-of-the-art performance, outperforming both traditional and learning-based methods with excellent generalization.

\begin{table*}
  \centering
  \renewcommand{\arraystretch}{1.05} 
      \captionsetup{labelfont={color=black}}
      \caption{Quantitative results on ETH3D dataset at threshold $2cm$ and $10cm$.}
    \resizebox{0.9\linewidth}{!}{
        \begin{tabular}{c|ccc|ccc|ccc|ccc}
        \hline
        \multirow{3}{*}{Method} & \multicolumn{6}{c|}{Test} & \multicolumn{6}{c}{Train} \\
        \cline{2-13} & \multicolumn{3}{c|}{$2cm$} & \multicolumn{3}{c|}{$10cm$} & \multicolumn{3}{c|}{$2cm$} & \multicolumn{3}{c}{$10cm$}  \\
        \cline{2-13} & F$_1$ & Comp. & Acc. & F$_1$ & Comp. & Acc. & F$_1$ & Comp. & Acc. & F$_1$ & Comp. & Acc. \\   
        \hline 
        \multirow{1}{*}{PatchMatchNet\cite{Iter-MVS}} & 73.12 & 77.46 & 69.71 & 91.91 & 92.05 & 91.98 & 64.21 & 65.43 & 64.81 & 85.70 & 83.28 & 89.98 \\   
        \multirow{1}{*}{IterMVS-LS\cite{Iter-MVS}} & 80.06 & 76.49 & 84.73 & 92.29 & 88.34 & 96.92 & 71.69 & 66.08 & 79.79 & 88.60 & 82.62 & 96.35 \\   
        \multirow{1}{*}{MVSTER\cite{MG-MVS}} & 79.01 & 82.47 & 77.09 & 93.20 & 92.71 & 94.21 & 72.06 & 76.92 & 68.08 & 91.73 & 91.91 & 91.97 \\  
        \multirow{1}{*}{EPP-MVSNet\cite{EPP-MVSNet}} & 83.40 & 81.79 & 85.47 & 95.22 & 93.75 & 96.84 & 74.00 & 67.58 & 82.76 & 92.13 & 87.72 & 97.29 \\  
        \multirow{1}{*}{EP-Net\cite{EP-Net}} & 83.72 & 87.84 & 80.37 & 95.20 & 96.82 & 93.72 & 79.08 & 79.28 & 79.36 & 93.92 & 93.69 & 94.33 \\         
        \hline 
        \multirow{1}{*}{TAPA-MVS\cite{TAPA-MVS}} & 79.15 & 74.94 & 85.71 & 92.30 & 90.35 & 94.93 & 77.69 & 71.45 & 85.88 & 93.69 & 90.98 & 96.79 \\
        \multirow{1}{*}{ACMP\cite{ACMP}} & 81.51 & 75.58 & \textbf{90.54} & 92.62 & 88.71 & 97.47 & 79.79 & 72.15 & \textbf{90.12} & 92.03 & 87.15 & 97.97 \\   
        \multirow{1}{*}{APD-MVS\cite{APD-MVS}} & 87.44 & 85.93 & 89.54 & 96.95 & 96.95 & 97.00 & 86.84 & 84.83 & 89.14 & 97.12 & 96.79 & 97.47 \\   
        \multirow{1}{*}{HPM-MVS++\cite{HPM-MVS}} & 89.02 & 89.37 & 88.93 & 97.34 & 97.72 & 96.99 & 87.09 & 85.64 & 88.74 & 97.22 & 96.91 & 97.56 \\  
        \multirow{1}{*}{SD-MVS\cite{SD-MVS} (base)} & 88.06 & 87.49 & 88.96 & 97.41 & 97.51 & 97.37 & 86.94 & 84.52 & 89.63 & 97.35 & 96.87 & 97.84 \\  
        \multirow{1}{*}{DVP-MVS \cite{DVP-MVS} (base)} & \textbf{89.60} & \textbf{89.41} & \textcolor{red}{\textbf{91.32}} & \textbf{97.91} & \textbf{98.07} & \textbf{97.78} & \textbf{88.67} & \textcolor{red}{\textbf{88.07}} & 89.40 & \textbf{97.93} & \textbf{97.61} & \textbf{98.27} \\  
        \hline 
        \multirow{1}{*}{DVP-MVS++ (ours)} & \textcolor{red}{\textbf{90.23}} & \textcolor{red}{\textbf{90.53}} & 90.07 & \textcolor{red}{\textbf{98.06}} & \textcolor{red}{\textbf{98.18}} & \textcolor{red}{\textbf{97.96}} & \textcolor{red}{\textbf{89.14}} & \textbf{87.98} & \textcolor{red}{\textbf{90.50}} & \textcolor{red}{\textbf{98.07}} & \textcolor{red}{\textbf{97.71}} & \textcolor{red}{\textbf{98.44}} \\  
        \hline
        \end{tabular}%
    }
    \label{table:ETH3D}%
    \vspace{-0.09in}
\end{table*}%

\begin{table*}
  \centering
  \renewcommand{\arraystretch}{1.05} 
      \captionsetup{labelfont={color=black}}
  \caption{Quantitative results on partial scenes of Strecha dataset (\emph{Fountain} and \emph{HerzJesu}) at threshold $2cm$ and $10cm$.}
    \resizebox{0.9\linewidth}{!}{
        \begin{tabular}{c|ccc|ccc|ccc|ccc}
        \hline
        \multirow{3}{*}{Method} & \multicolumn{6}{c|}{Fountain} & \multicolumn{6}{c}{HerzJesu} \\
        \cline{2-13} & \multicolumn{3}{c|}{$2cm$} & \multicolumn{3}{c|}{$10cm$} & \multicolumn{3}{c|}{$2cm$} & \multicolumn{3}{c}{$10cm$}  \\
        \cline{2-13} & F$_1$ & Comp. & Acc. & F$_1$ & Comp. & Acc. & F$_1$ & Comp. & Acc. & F$_1$ & Comp. & Acc. \\   
        \hline 
        \multirow{1}{*}{OpenMVS\cite{OpenMVS}} & 74.77 & 70.47 & 79.62 & 87.37 & 83.49 & 91.63 & 69.67 & 61.85 & 79.76 & 79.94 & 70.37 & 92.53 \\
        \multirow{1}{*}{PatchMatchNet\cite{Iter-MVS}} & 69.06 & 68.73 & 69.4 & 84.61 & 82.13 & 87.25 & 62.43 & 59.32 & 65.89 & 77.11 & 69.25 & 86.98 \\  
        \multirow{1}{*}{IterMVS-LS\cite{Iter-MVS}} & 75.63 & 69.45 & 83.02 & 87.78 & 82.61 & 93.65 & 69.73 & 60.14 & 82.97 & 80.33 & 69.83 & 94.55 \\
        \multirow{1}{*}{MVSTER\cite{Iter-MVS}} & 76.05 & 78.62 & 73.65 & 89.72 & 90.85 & 88.61 & 71.04 & 71.43 & 70.65 & 83.26 & 79.04 & 87.95 \\   
        \hline 
        \multirow{1}{*}{ACMM\cite{ACMM}} & 75.48 & 67.32 & 85.89 & 87.28 & 81.29 & 94.23 & 68.24 & 57.56 & 83.78 & 79.11 & 67.85 & 94.86 \\
        \multirow{1}{*}{ACMMP\cite{ACMMP}} & 80.47 & 73.75 & \textcolor{red}{\textbf{88.53}} & 89.88 & 84.36 & \textcolor{red}{\textbf{96.18}} & 73.13 & 63.41 & \textcolor{red}{\textbf{86.38}} & 83.01 & 72.65 & \textcolor{red}{\textbf{96.81}} \\
        \multirow{1}{*}{APD-MVS\cite{APD-MVS}} & 83.71 & 80.63 & 87.04 & 91.60 & 87.89 & 95.64 & 77.89 & 72.13 & 84.65 & 85.63 & 77.18 & 96.16 \\  
        \multirow{1}{*}{HPM-MVS++\cite{HPM-MVS}} & 84.02 & 81.84 & 86.32 & 91.73 & 88.03 & 95.75 & 78.41 & 73.69 & 83.77 & 85.78 & 77.34 & 96.28 \\  
        \multirow{1}{*}{SD-MVS\cite{SD-MVS}} & 83.78 & 80.32 & 87.56 & 91.70 & 87.92 & 95.85 & 77.96 & 71.84 & 85.23 & 85.82 & 77.26 & 96.51 \\  
        \multirow{1}{*}{DVP-MVS \cite{DVP-MVS} (base)} & \textbf{85.07} & \textbf{82.81} & 87.45 & \textbf{92.07} & \textbf{88.62} & 95.81 & \textbf{79.70} & \textbf{74.92} & 85.13 & \textbf{86.29} & \textbf{78.06} & 96.46 \\  
        \hline 
        \multirow{1}{*}{DVP-MVS++ (ours)} & \textcolor{red}{\textbf{86.38}} & \textcolor{red}{\textbf{84.85}} & \textbf{87.96} & \textcolor{red}{\textbf{92.46}} & \textcolor{red}{\textbf{89.19}} & \textbf{95.98} & \textcolor{red}{\textbf{81.34}} & \textcolor{red}{\textbf{77.32}} & \textbf{85.79} & \textcolor{red}{\textbf{86.75}} & \textcolor{red}{\textbf{78.65}} & \textbf{96.72} \\  
        \hline 
        \end{tabular}%
    }
  \label{table:strecha}%
  \vspace{-0.09in}
\end{table*}%

\begin{table}
  \centering
  \renewcommand{\arraystretch}{1.07} 
    \captionsetup{labelfont={color=black}}
    \caption{Quantitative results on TNT dataset at given threshold.}
    \resizebox{\linewidth}{!}{
        \begin{tabular}{c|ccc|ccc}
        \hline
        \multirow{2}{*}{Method} & \multicolumn{3}{c|}{Intermediate} & \multicolumn{3}{c}{Advanced} \\
        \cline{2-7} & F$_1$ & Rec. & Pre. & F$_1$ & Rec. & Pre. \\   
        \hline 
        \multirow{1}{*}{PatchMatchNet\cite{PatchMatchNet}} & 53.15 & 69.37 & 43.64 & 32.31 & 41.66 & 27.27 \\ 
        \multirow{1}{*}{IterMVS-LS\cite{Iter-MVS}} & 56.94 & 74.69 & 47.53 & 34.17 & 44.19 & 28.70 \\  
        \multirow{1}{*}{AGG-CVCNet\cite{AGG-CVCNet}} & 57.81 & 71.71 & 49.04 & 28.96 & 28.28 & 35.33 \\  
        \multirow{1}{*}{MVSTER\cite{MVSTER}} & 60.92 & 77.50 & 50.17 & 37.53 & 45.90 & 33.23 \\  
        \multirow{1}{*}{EPP-MVSNet\cite{EPP-MVSNet}} & 61.68 & 75.58 & 53.09 & 35.72 & 34.63 & \textbf{40.09} \\
        \multirow{1}{*}{EP-Net\cite{EP-Net}} & 63.68 & 72.57 & 57.01 & \textbf{40.52} & \textbf{50.54} & 34.26 \\ 
        \hline 
        \multirow{1}{*}{PCF-MVS\cite{PCF-MVS}} & 53.39 & 58.85 & 50.04 & 34.59 & 34.35 & 35.84 \\  
        \multirow{1}{*}{ACMMP\cite{ACMMP}} & 59.38 & 68.50 & 53.28 & 37.84 & 44.64 & 33.79 \\  
        \multirow{1}{*}{HPM-MVS++\cite{HPM-MVS}} & 61.59 & 73.79 & 54.01 & 39.65 & 41.09 &  \textcolor{red}{\textbf{40.79}} \\  
        \multirow{1}{*}{APD-MVS\cite{APD-MVS}} & 63.64 & 75.06 & 55.58 & 39.91 & 49.41 & 33.77 \\  
        \multirow{1}{*}{SD-MVS\cite{SD-MVS}} & 63.31 & 76.63 & 53.78 & 40.18 & 50.31 & 33.81 \\ 
        \multirow{1}{*}{DVP-MVS\cite{DVP-MVS}} & \textbf{64.76} &  \textcolor{red}{\textbf{78.69}} & 55.04 & 40.23 &  \textcolor{red}{\textbf{54.21}} & 32.16 \\  
        \hline 
        \multirow{1}{*}{DVP-MVS++ (ours)} &  \textcolor{red}{\textbf{65.95}} & \textbf{78.11} &  \textcolor{red}{\textbf{57.41}} &  \textcolor{red}{\textbf{41.65}} & \textbf{50.54} & 36.21 \\  
        \hline
        \end{tabular}%
    }
  \label{table:TNT}%
  \vspace{-0.09in}
\end{table}%

\begin{figure*}
\centering
\includegraphics[width=\linewidth]{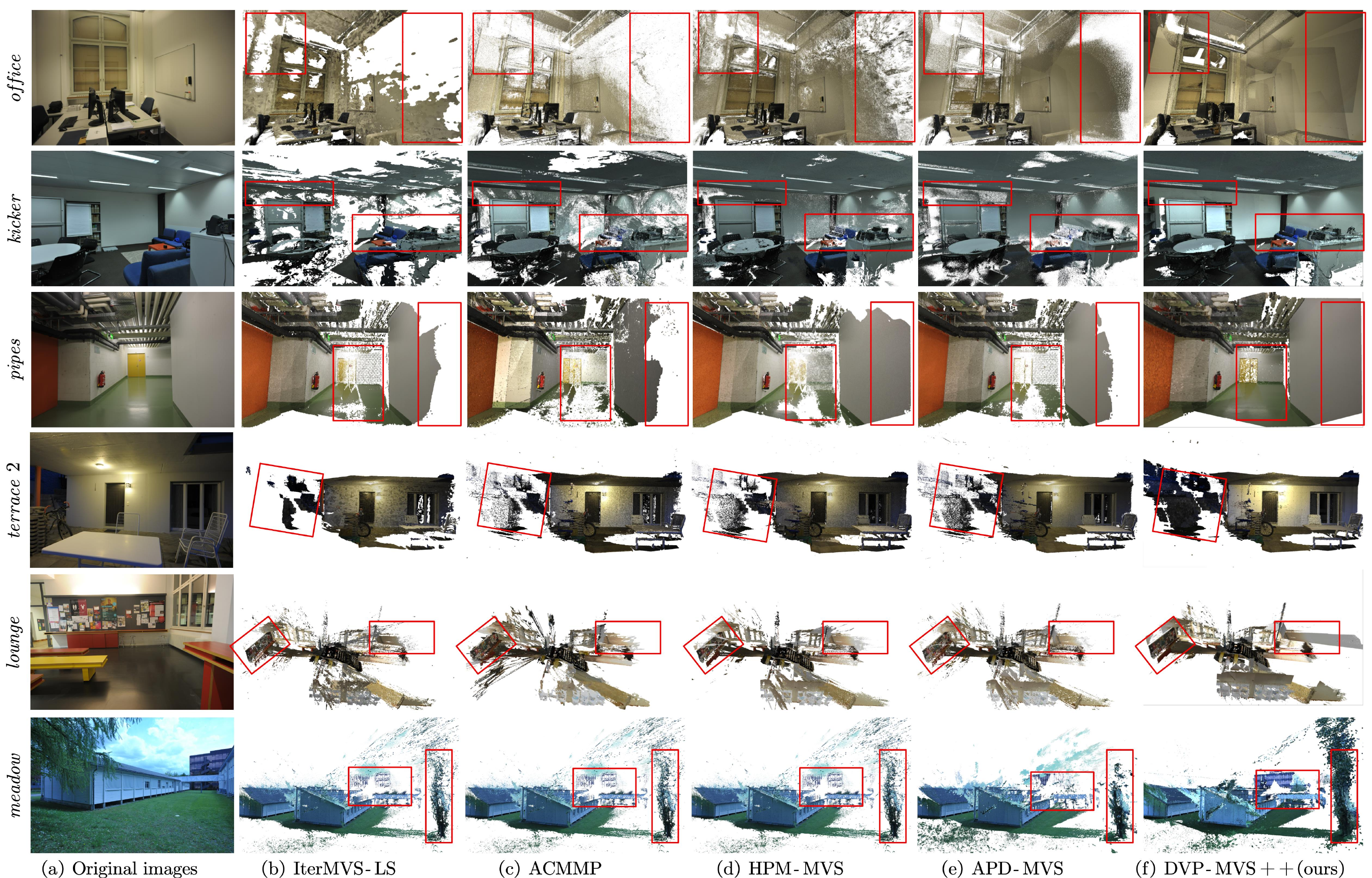}
\caption{
Visualized point cloud results between different methods on partial scenes of ETH3D datasets (\emph{office}, \emph{kicker}, \emph{pipes}, \emph{terrace 2}, \emph{lounge} and \emph{meadow}). Obviously, our method can effectively reconstruct textureless areas such as floors and walls without detail distortion. 
}
\label{fig: ETH3D}
\end{figure*}

\begin{figure*}
\centering
\includegraphics[width=\linewidth]{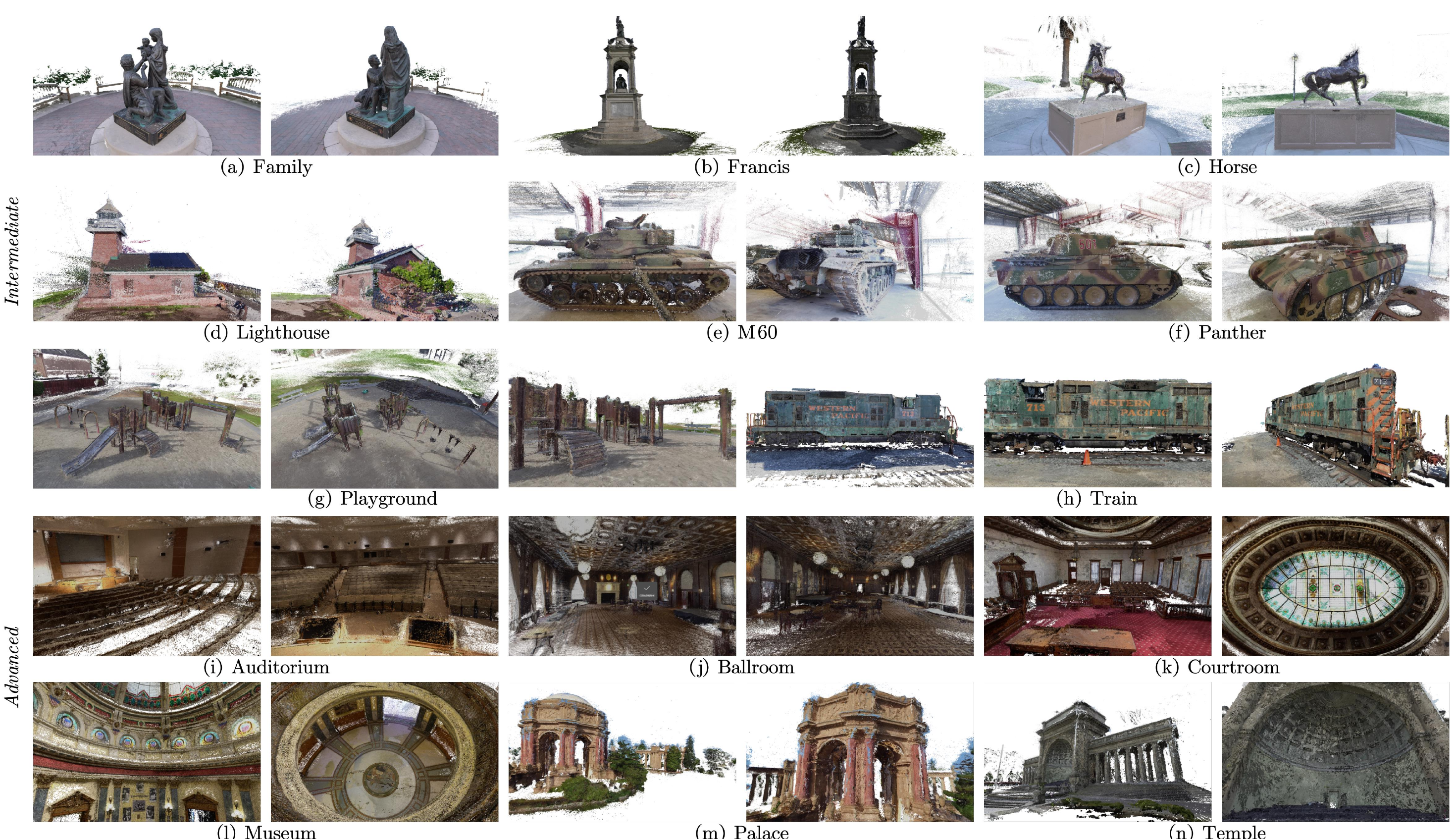}
\caption{
Reconstructed point clouds of our method on Tanks \& Temples dataset without any fine-tuning.
}
\vspace{-0.1in}
\label{fig: TNT}
\end{figure*}

\subsection{Datasets and Implementation Details}
The ETH3D high-resolution benchmark \cite{ETH3D} consists of 25 scenes, each with images at a resolution of 6,221 × 4,146. This dataset poses significant challenges for reconstruction due to its wide range of viewpoints and diverse scene types. It is divided into training and testing sets, with the training set containing 13 scenes, which come with both ground truth point clouds and depth maps. The testing set, consisting of 12 scenes, retains its ground truth point clouds and depth maps on the official benchmark platform.

The Tanks and Temples (TNT) Benchmark \cite{TNT} includes 14 distinct scenes with a resolution of 1,920 × 1,080, featuring individual objects such as the francis and panther tanks, as well as large indoor scenes like museums and ballrooms. The ground truth depth maps for this dataset are captured using high-precision industrial laser scanners. Both the ETH3D and TNT datasets are split into training and testing subsets, with our results available on their respective official websites.


The Strecha dataset \cite{strecha} contains six outdoor scenes at a resolution of 3,072 × 2,048. Among these, only the Fountain and HerzJesu scenes provide ground truth point clouds for evaluation, consisting of 11 and 8 images, respectively.

To achieve faster processing times, we downsample the images to half their original resolution in the ETH3D and Strecha datasets, while using the full resolution in the TNT datasets. This adjustment increases runtime efficiency without sacrificing the robustness of our method.

Our method is implemented on a machine whose configuration contains an Intel(R) Core(TM) i7-10700 CPU @ 2.90GHz and an NVIDIA GeForce RTX 3080 graphics card. 
Concerning parameter setting, $\{\eta, \varphi, \phi, \gamma, \kappa, \delta, \varepsilon, \alpha, \beta, \mu\} = \{3 \times 10^{2}, 0.5, 0.4, 1.2, 0.7, 0.8, 2, 1, 4, 3\}$. 

\subsection{Point Cloud Evaluation}
To assess the effectiveness of our approach, we evaluate the accuracy (Acc.), completeness (Comp.), and F1 score of the reconstructed point clouds and compare them with other methods. Specifically, we compare our method against traditional multi-view stereo (MVS) algorithms such as ACMM \cite{ACMM}, ACMP \cite{ACMP}, ACMMP \cite{ACMMP}, APD-MVS \cite{APD-MVS}, HPM-MVS++ \cite{HPM-MVS}, and SD-MVS \cite{SD-MVS}, as well as learning-based MVS methods including PatchMatchNet \cite{PatchMatchNet}, IterMVS-LS \cite{Iter-MVS}, MVSTER \cite{MVSTER}, EPP-MVSNet \cite{EPP-MVSNet}, and EP-Net \cite{EP-Net}.
Quantitative results on the ETH3D and TNT datasets are reported in Table \ref{table:ETH3D} and Table \ref{table:TNT}, respectively. The best-performing results are highlighted in bold red, while the second-best are marked in bold black. For fairness and consistency, the results of other methods are sourced directly from their official websites.

As presented in the tables above, our method achieves the highest F1 scores and completeness on both the training and testing sets of the ETH3D dataset, confirming its superior effectiveness. For the TNT dataset, we evaluate our method without any fine-tuning to demonstrate its generalization capability. On both the intermediate and the advanced set, our method achieves the highest F1 score and completeness, further illustrating the robustness of our approach.

Qualitative results for both the ETH3D and the TNT datasets are shown in Fig. \ref{fig: ETH3D} and Fig. \ref{fig: TNT}, respectively. As seen in Fig. \ref{fig: ETH3D}, our method delivers the most complete and visually realistic point clouds, particularly in low-texture regions such as walls and floors (marked by red boxes), and it avoids noticeable distortion in fine details. Fig. \ref{fig: TNT} further supports this observation, demonstrating our method’s ability to faithfully reconstruct both textureless and high-detail areas. Additional qualitative examples on both the ETH3D and the Strecha dataset, along with comprehensive memory usage and runtime comparisons, are available in the supplementary materials.

\begin{table}
    \centering
    \renewcommand{\arraystretch}{1.05} 
    \caption{Quantitative results of the ablation studies on ETH3D benchmark to validate each proposed component.}
    \resizebox{\linewidth}{!}{
        \begin{tabular}{c|ccc|ccc}
        \hline
        \multirow{2}{*}{Method} & \multicolumn{3}{c|}{$2cm$} & \multicolumn{3}{c}{$10cm$} \\
        \cline{2-7} & F$_1$ & Comp. & Acc. & F$_1$ & Comp. & Acc. \\   
        \hline  
        \multirow{1}{*}{w/o. Agn.} & 87.25 & 85.43 & 89.31 & 96.78 & 95.94 & 97.63 \\ 
        \multirow{1}{*}{w/o. Mon.} & 88.43 & 87.23 & 89.82 & 97.57 & 97.06 & 98.09 \\ 
        \multirow{1}{*}{w/o. Ero.} & 88.36 & 87.14 & 89.76 & 97.51 & 96.98 & 98.05 \\ 
        \multirow{1}{*}{w/o. Fil.} & 88.93 & 87.69 & 90.35 & 97.89 & 97.42 & 98.37 \\ 
        \hline
        \multirow{1}{*}{w/o. Har.} & 87.48 & 85.96 & 89.23 & 96.93 & 96.31 & 97.58 \\ 
        \multirow{1}{*}{w/o. Max.} & 88.65 & 87.56 & 89.91 & 97.70 & 97.27 & 98.14 \\ 
        \multirow{1}{*}{w/o. Res.} & 88.57 & 87.45 & 89.86 & 97.65 & 97.20 & 98.11 \\ 
        \multirow{1}{*}{w/o. Vis.} & 88.32 & 87.09 & 89.71 & 97.48 & 96.91 & 98.05 \\ 
        \hline
        \multirow{1}{*}{w/o. Geo.} & 87.80 & 86.63 & 89.16 & 97.14 & 96.75 & 97.54 \\ 
        \multirow{1}{*}{w/o. Hig.} & 88.74 & 87.75 & 89.90 & 97.77 & 97.61 & 97.93 \\ 
        \multirow{1}{*}{w/o. Nor.} & 88.41 & 87.52 & 89.48 & 97.55 & 97.38 & 97.72 \\ 
        \multirow{1}{*}{w/o. Dep.} & 88.59 & 87.61 & 89.76 & 97.67 & 97.50 & 97.85 \\ 
        \hline
        \multirow{1}{*}{DVP-MVS++} & \textcolor{red}{\textbf{89.14}} & \textcolor{red}{\textbf{87.98}} & \textcolor{red}{\textbf{90.50}} & \textcolor{red}{\textbf{98.07}} & \textcolor{red}{\textbf{97.71}} & \textcolor{red}{\textbf{98.44}} \\  
        \hline
        \end{tabular}%
    }
    \label{table: ablation study}%
\end{table}%

\subsection{Ablation Study}
To validate the devised modules of our method, we conduct a series of experiments for ablations study to illustrate the effectiveness of each proposed module, as shown in Tab. \ref{table: ablation study}. 
\subsubsection{Depth-Normal-Edge Aligned Prior} 
We separately remove the whole depth-normal-edge aligned prior (w/o. Agn.), monocular normal map (w/o. Mon.), the erosion-dilation strategy (w/o. Ero.) and pixel-wise filtering (w/o. Fil.). 
Notably, w/o Agn. yields the lowest F$_1$ score, underscoring the importance of integrating both monocular depth and normal map for the extraction of depth boundaries for patch deformation. Since w/o Mon. variant excludes monocular normals, it essentially replicates the depth-edge aligned prior proposed in DVP-MVS. The F$_1$ scores of DVP-MVS++ outperforms w/o Mon., indicates that the depth-normal-edge aligned prior significantly outperforms the depth-edge aligned prior, confirming the effectiveness of monocular normal maps in generating fine-grained depth edges.
Moreover, the F$_1$ score of w/o Ero. surpasses w/o Fil., demonstrating that erosion-dilation strategy exerts is more critical than pixel-wise filtering. 

\subsubsection{Harmonized Cross-View Prior}
We individually exclude the entire harmonized cross-view prior (w/o. Har.), area-maximization strategy (w/o. Max.), visibility map restoration (w/o. Res.) and visibility-aware patch deformation (w/o. Vis.).
Notably, w/o Har. produces the lowest F$_1$ score, underlining the importance of the harmonized cross-view prior for effective patch deformation. 
Since w/o Max. lacks area-maximization strategy, it essentially represents the cross-view prior proposed in DVP-MVS. The F$_1$ scores of DVP-MVS++ outperforms w/o Max., meaning the area-maximization strategy can achieve a more balanced coverage of deformed patches within homogeneous areas.
Additionally, w/o Res. achieves higher F$_1$ scores than w/o Vis., emphasizing that visibility-aware patch deformation has a greater impact than visibility map restoration.

\yzl{\subsubsection{Geometry Propagation and Refinement}
We respectively exclude the whole highlight geometry-driven module (w/o. Geo.), the SHIQ for highlight correction (w/o. Hig.), the normal constraint (w/o. Nor.) and depth constraint (w/o. Dep.) during both propagation and refinement stages.
Evidently, excluding the highlight geometric constraint (w/o Geo.) yields the lowest F$_1$ score, underscoring its importance. The configuration without SHIQ highlight correction (w/o Hig.) effectively reverts to the geometric constraint from DVP-MVS.} The F$_1$ scores of DVP-MVS++ outperforms w/o Hig., indicating the effectiveness of employing SHIQ for highlight correction.
Furthermore, w/o Nor. shows a lower F$_1$ score than w/o Dep., highlighting that hemispherical normal aggregation is more influential than epipolar depth aggregation during propagation and refinement.

\subsubsection{Ablation Studies on Key Parameters}
To further justify our choice of parameters, we conduct ablation studies on key parameters, as shown in Table \ref{table: parameters}. The best, second-best F$_1$ scores alongside our selected parameters are respectively highlighted in bold, red, and blue.
As shown in the table, the parameters we selected generally correspond to the best reconstruction results. Except for critical parameters like $\eta$, $\varphi$, $\phi$, and $\gamma$, most other parameters exhibit low sensitivity.

For the Depth-Normal-Edge Aligned Prior, although the F$_1$ score improves as $\eta$ decreases, it also leads to longer runtimes. Therefore, we choose $\eta = 5 \times 10^{2}$ as a balance between time efficiency and performance. 
Moreover, the difference in F$_1$ score for $\gamma$ between 1.2 and 1.6 is negligible. 
Additionally, the F$_1$ score is slightly higher when $\kappa = 0.6$, but the improvement is minimal compared with the F$_1$ score generated by $\kappa = 0.8$. 

For the Harmonized Cross-View Prior, the best results are obtained when $\varepsilon = 2$, with a noticeable drop in F1 scores when it is smaller. \yzl{Similarly, for the geometry propagation and refinement section, the best result is achieved when $\mu = 3$, with a sharp drop in F1 scores when reducing $\mu$.}

\begin{table}
    \centering
    \renewcommand{\arraystretch}{1.07} 
    \caption{Quantitative evaluation of hyperparameter variability on the F$_1$ Score at threshold $2cm$ of ETH3D training dataset.}
    \resizebox{0.95\linewidth}{!}{ 
        \begin{tabular}{c|c|c|c|c|c}
        \hline  
        \multirow{1}{*}{Par.} & \multicolumn{5}{c}{Depth-Normal-Edge Aligned Prior} \\
        \hline  
        \multirow{1}{*}{$\eta$} & 1$\times$10$^{2}$ & 2$\times$10$^{2}$ & \textcolor{blue}{\textbf{5$\times$10$^{2}$}} & 6$\times$10$^{2}$ & 9$\times$10$^{2}$ \\ 
        \hline  
        \multirow{1}{*}{F$_1$} & \textcolor{red}{\textbf{89.21}} & \textbf{89.18} & 89.14 & 88.96 & 88.78 \\ 
        \hline  
        \multirow{1}{*}{$\varphi$} & 0.1 & 0.3 & \textcolor{blue}{\textbf{0.5}} & 0.7 & 0.9 \\ 
        \hline  
        \multirow{1}{*}{F$_1$} & 88.75 & \textbf{89.01} & \textcolor{red}{\textbf{89.14}} & 88.92 & 88.63 \\ 
        \hline  
        \multirow{1}{*}{$\phi$} & 0.1 & 0.2 & \textcolor{blue}{\textbf{0.4}} & 0.6 & 0.8 \\ 
        \hline  
        \multirow{1}{*}{F$_1$} & 88.8 & \textbf{89.03} & \textcolor{red}{\textbf{89.14}} & 88.98 & 88.76 \\ 
        \hline  
        \multirow{1}{*}{$\gamma$} & 0.4 & 0.8 & \textcolor{blue}{\textbf{1.2}} & 1.6 & 2.0 \\ 
        \hline  
        \multirow{1}{*}{F$_1$} & 88.73 & 89 & \textcolor{red}{\textbf{89.14}} & \textcolor{red}{\textbf{89.14}} & \textbf{89.03} \\ 
        \hline  
        \multirow{1}{*}{$\kappa$} & 0.5 & 0.6 & \textcolor{blue}{\textbf{0.7}} & 0.8 & 0.9 \\ 
        \hline  
        \multirow{1}{*}{F$_1$} & 89.12 & \textcolor{red}{\textbf{89.16}} & \textbf{89.14} & 89.1 & 89.05 \\ 
        \hline  
        \multirow{1}{*}{$\delta$} & 0.4 & 0.6 & \textcolor{blue}{\textbf{0.8}} & 1.2 & 1.6 \\ 
        \hline  
        \multirow{1}{*}{F$_1$} & 88.92 & \textbf{89.06} & \textcolor{red}{\textbf{89.14}} & 89.02 & 88.81 \\ 
        \hline  
        \multirow{1}{*}{Par.} & \multicolumn{5}{c}{Harmonized Cross-View Prior} \\
        \hline  
        \multirow{1}{*}{$\varepsilon$} & 1.0 & 1.5 & \textcolor{blue}{\textbf{2.0}} & 3.0 & 4.0 \\ 
        \hline  
        \multirow{1}{*}{F$_1$} & 88.81 & 89.02 & \textcolor{red}{\textbf{89.14}} & \textbf{89.05} & 88.87 \\ 
        \hline  
        \multirow{1}{*}{Par.} & \multicolumn{5}{c}{Geometry Prop. and Refine.} \\
        \hline  
        \multirow{1}{*}{$\mu$} & 1 & 2 & \textcolor{blue}{\textbf{3}} & 4 & 5 \\ 
        \hline  
        \multirow{1}{*}{F$_1$} & 88.85 & 89.03 & \textcolor{red}{\textbf{89.14}} & \textbf{89.07} & 88.95 \\ 
        \hline  
        \end{tabular}%
    }
    \label{table: parameters}%
    \vspace{-0.05in}
\end{table}%

\section{Conclusion} 
In this work, we presented DVP-MVS++, a novel method that synergizes both depth-normal-edge aligned and harmonized cross-view prior to address key challenges in patch deformation. By integrating monocular depth, normal estimation, and edge detection, our approach generates fine-grained homogeneous boundaries to prevent edge-skipping and ensure robust patch deformation. Moreover, we construct visibility maps and further restore them through depth reprojection and area-maximization strategy, thus obtain harmonized cross-view prior for visibility-aware patch deformation. 
\yzl{Additionally, we employ SHIQ for highlight correction to facilitate geometry consistency for propagation and refinement stages.} Experiments on ETH3D, TNT and Strecha benchmarks demonstrate our SOTA performance and generalization. 

\bibliography{DVP-MVS++}

\begin{thebibliography}{10}
\providecommand{\url}[1]{#1}
\csname url@samestyle\endcsname
\providecommand{\newblock}{\relax}
\providecommand{\bibinfo}[2]{#2}
\providecommand{\BIBentrySTDinterwordspacing}{\spaceskip=0pt\relax}
\providecommand{\BIBentryALTinterwordstretchfactor}{4}
\providecommand{\BIBentryALTinterwordspacing}{\spaceskip=\fontdimen2\font plus
\BIBentryALTinterwordstretchfactor\fontdimen3\font minus \fontdimen4\font\relax}
\providecommand{\BIBforeignlanguage}[2]{{%
\expandafter\ifx\csname l@#1\endcsname\relax
\typeout{** WARNING: IEEEtran.bst: No hyphenation pattern has been}%
\typeout{** loaded for the language `#1'. Using the pattern for}%
\typeout{** the default language instead.}%
\else
\language=\csname l@#1\endcsname
\fi
#2}}
\providecommand{\BIBdecl}{\relax}
\BIBdecl

\bibitem{Augmented}
M.~Cao, L.~Zheng, W.~Jia, H.~Lu, and X.~Liu, ``Accurate 3-{{D}} reconstruction under {{IoT}} environments and its applications to augmented reality,'' \emph{IEEE Transactions on Industrial Informatics}, vol.~17, no.~3, pp. 2090--2100, 2020.

\bibitem{Autonomous}
Y.~Wei, L.~Zhao, W.~Zheng, Z.~Zhu, J.~Zhou, and J.~Lu, ``Surroundocc: {{Multi-camera}} 3d occupancy prediction for autonomous driving,'' in \emph{Proceedings of the {{IEEE}}/{{CVF International Conference}} on {{Computer Vision}}}, 2023, pp. 21\,729--21\,740.

\bibitem{3D-printed}
Z.~Zhu, D.~W.~H. Ng, H.~S. Park, and M.~C. McAlpine, ``{{3D-printed}} multifunctional materials enabled by artificial-intelligence-assisted fabrication technologies,'' \emph{Nature Reviews Materials}, vol.~6, no.~1, pp. 27--47, 2021.

\bibitem{SED-MVS}
Z.~Yuan, Z.~Yang, Y.~Cai, K.~Wu, M.~Liu, D.~Zhang, H.~Jiang, Z.~Li, and Z.~Wang, ``{{SED-MVS}}: {{Segmentation-Driven}} and {{Edge-Aligned Deformation Multi-View Stereo}} with {{Depth Restoration}} and {{Occlusion Constraint}},'' Mar. 2025.

\bibitem{MSP-MVS}
Z.~Yuan, C.~Liu, F.~Shen, Z.~Li, J.~Luo, T.~Mao, and Z.~Wang, ``{{MSP-MVS}}: {{Multi-Granularity Segmentation Prior Guided Multi-View Stereo}},'' Dec. 2024.

\bibitem{wang2024efficient}
S.~Wang, B.~Li, and Y.~Dai, ``Efficient multi-view stereo by dynamic cost volume and cross-scale propagation,'' \emph{IEEE Transactions on Circuits and Systems for Video Technology}, vol.~34, no.~10, pp. 9414--9427, 2024.

\bibitem{fang2020ugnet}
Y.~Fang, W.~Zhu, and Q.~Zhu, ``Ugnet: Underexposed images enhancement network based on global illumination estimation,'' in \emph{2020 IEEE International Conference on Visual Communications and Image Processing (VCIP)}.\hskip 1em plus 0.5em minus 0.4em\relax IEEE, 2020, pp. 415--418.

\bibitem{chen2024dsc3d}
J.~Chen, Q.~Song, W.~Guo, and R.~Huang, ``Dsc3d: Deformable sampling constraints in stereo 3d object detection for autonomous driving,'' \emph{IEEE Transactions on Circuits and Systems for Video Technology}, 2024.

\bibitem{li2024focus}
H.~Li, D.-H. Zhai, Q.~Liu, K.~Tian, Y.~Yang, Z.~Chang, S.~Wang, and Y.~Xia, ``Focus-transunet3d: High-precision model for 3d segmentation of medical point targets,'' \emph{IEEE Transactions on Circuits and Systems for Video Technology}, 2024.

\bibitem{ye2024self}
X.~Ye, Y.~Ou, B.~Wu, R.~Xu, and H.~Li, ``Self-supervised monocular depth estimation from videos via adaptive reconstruction constraints,'' \emph{IEEE Transactions on Circuits and Systems for Video Technology}, 2024.

\bibitem{ETH3D}
T.~Schops, J.~L. Schonberger, S.~Galliani, T.~Sattler, K.~Schindler, M.~Pollefeys, and A.~Geiger, ``A multi-view stereo benchmark with high-resolution images and multi-camera videos,'' in \emph{Proc. IEEE/CVF Conf. Comput. Vis. Pattern Recognit. (CVPR)}, July 2017.

\bibitem{TNT}
A.~Knapitsch, J.~Park, Q.-Y. Zhou, and V.~Koltun, ``Tanks and temples: Benchmarking large-scale scene reconstruction,'' 2017.

\bibitem{strecha}
C.~Strecha, W.~von Hansen, L.~Van~Gool, P.~Fua, and U.~Thoennessen, ``On benchmarking camera calibration and multi-view stereo for high resolution imagery,'' in \emph{Proc. IEEE/CVF Conf. Comput. Vis. Pattern Recognit. (CVPR)}, 2008, pp. 1--8.

\bibitem{Blendedmvs}
Y.~Yao, Z.~Luo, S.~Li, J.~Zhang, Y.~Ren, L.~Zhou, T.~Fang, and L.~Quan, ``Blendedmvs: {{A}} large-scale dataset for generalized multi-view stereo networks,'' in \emph{Proceedings of the {{IEEE}}/{{CVF Conference}} on {{Computer Vision}} and {{Pattern Recognition}}}, 2020, pp. 1790--1799.

\bibitem{PM}
C.~Barnes, E.~Shechtman, A.~Finkelstein, and D.~B. Goldman, ``Patchmatch: A randomized correspondence algorithm for structural image editing,'' \emph{ACM Trans. Graph.}, p.~24, 2009.

\bibitem{TSAR-MVS}
Z.~Yuan, J.~Cao, Z.~Wang, and Z.~Li, ``Tsar-mvs: {{Textureless-aware}} segmentation and correlative refinement guided multi-view stereo,'' \emph{Pattern Recognition}, p. 110565, 2024.

\bibitem{ACMMP}
Q.~Xu, W.~Kong, W.~Tao, and M.~Pollefeys, ``Multi-{{Scale Geometric Consistency Guided}} and {{Planar Prior Assisted Multi-View Stereo}},'' \emph{IEEE Trans. Pattern Anal. Mach. Intell.}, pp. 1--18, 2022.

\bibitem{HPM-MVS}
C.~Ren, Q.~Xu, S.~Zhang, and J.~Yang, ``Hierarchical prior mining for non-local multi-view stereo,'' in \emph{Proceedings of the {{IEEE}}/{{CVF International Conference}} on {{Computer Vision}}}, 2023, pp. 3611--3620.

\bibitem{PHI-MVS}
S.~Sun, Y.~Zheng, X.~Shi, Z.~Xu, and Y.~Liu, ``Phi-mvs: {{Plane}} hypothesis inference multi-view stereo for large-scale scene reconstruction,'' \emph{arXiv preprint arXiv:2104.06165}, 2021.

\bibitem{APD-MVS}
Y.~Wang, Z.~Zeng, T.~Guan, W.~Yang, Z.~Chen, W.~Liu, L.~Xu, and Y.~Luo, ``Adaptive patch deformation for textureless-resilient multi-view stereo,'' in \emph{Proc. IEEE/CVF Conf. Comput. Vis. Pattern Recognit. (CVPR)}, 2023, pp. 1621--1630.

\bibitem{depany2}
L.~Yang, B.~Kang, Z.~Huang, Z.~Zhao, X.~Xu, J.~Feng, and H.~Zhao, ``Depth {{Anything V2}},'' Jun. 2024.

\bibitem{Metric3D}
M.~Hu, W.~Yin, C.~Zhang, Z.~Cai, X.~Long, H.~Chen, K.~Wang, G.~Yu, C.~Shen, and S.~Shen, ``Metric3d v2: {{A}} versatile monocular geometric foundation model for zero-shot metric depth and surface normal estimation,'' \emph{IEEE Transactions on Pattern Analysis and Machine Intelligence}, 2024.

\bibitem{DVP-MVS}
Z.~Yuan, J.~Luo, F.~Shen, Z.~Li, C.~Liu, T.~Mao, and Z.~Wang, ``{{DVP-MVS}}: {{Synergize Depth-Edge}} and {{Visibility Prior}} for {{Multi-View Stereo}},'' Dec. 2024.

\bibitem{PMS}
M.~Bleyer, C.~Rhemann, and C.~Rother, ``Patchmatch stereo - stereo matching with slanted support windows,'' in \emph{British Mach. Vis. Conf. (BMVC)}, J.~Hoey, S.~J. McKenna, and E.~Trucco, Eds., September 2011, pp. 1--11.

\bibitem{Gipuma}
S.~Galliani, K.~Lasinger, and K.~Schindler, ``Massively parallel multiview stereopsis by surface normal diffusion,'' in \emph{Proc. IEEE/CVF Int. Conf. Comput. Vis. (ICCV)}, December 2015.

\bibitem{ACMM}
Q.~Xu and W.~Tao, ``Multi-scale geometric consistency guided multi-view stereo,'' in \emph{Proc. IEEE/CVF Conf. Comput. Vis. Pattern Recognit. (CVPR)}, June 2019.

\bibitem{MG-MVS}
Y.~Wang, T.~Guan, Z.~Chen, Y.~Luo, K.~Luo, and L.~Ju, ``Mesh-guided multi-view stereo with pyramid architecture,'' in \emph{Proc. IEEE/CVF Conf. Comput. Vis. Pattern Recognit. (CVPR)}, June 2020, pp. 2036--2045.

\bibitem{Pyramid}
J.~Liao, Y.~Fu, Q.~Yan, and C.~Xiao, ``Pyramid {{Multi}}-{{View Stereo}} with {{Local Consistency}},'' \emph{Computer Graphics Forum}, vol.~38, no.~7, pp. 335--346, Oct. 2019.

\bibitem{API-MVS}
S.~Sun, J.~Liu, Y.~Li, H.~Ying, Z.~Zhai, and Y.~Mou, ``Adaptive pixelwise inference multi-view stereo,'' in \emph{Thirteenth {{International Conference}} on {{Graphics}} and {{Image Processing}} ({{ICGIP}} 2021)}, D.~Xu and L.~Xiao, Eds.\hskip 1em plus 0.5em minus 0.4em\relax Kunming, China: SPIE, Feb. 2022, p.~77.

\bibitem{SD-MVS}
Z.~Yuan, J.~Cao, Z.~Li, H.~Jiang, and Z.~Wang, ``{SD-MVS:} segmentation-driven deformation multi-view stereo with spherical refinement and {EM} optimization,'' \emph{CoRR}, vol. abs/2401.06385, 2024.

\bibitem{MVSNet}
Y.~Yao, Z.~Luo, S.~Li, T.~Fang, and L.~Quan, ``Mvsnet: Depth inference for unstructured multi-view stereo,'' in \emph{Proc. Eur. Conf. Comput. Vis. (ECCV)}, September 2018.

\bibitem{R-MVSNet}
Y.~Yao, Z.~Luo, S.~Li, T.~Shen, T.~Fang, and L.~Quan, ``Recurrent mvsnet for high-resolution multi-view stereo depth inference,'' in \emph{Proceedings of the {{IEEE}}/{{CVF}} Conference on Computer Vision and Pattern Recognition}, 2019, pp. 5525--5534.

\bibitem{Iter-MVS}
F.~Wang, S.~Galliani, C.~Vogel, and M.~Pollefeys, ``Itermvs: Iterative probability estimation for efficient multi-view stereo,'' in \emph{Proc. IEEE/CVF Conf. Comput. Vis. Pattern Recognit. (CVPR)}, 2022, pp. 8606--8615.

\bibitem{Cas-MVSNet}
J.~Yang, W.~Mao, J.~M. Alvarez, and M.~Liu, ``Cost volume pyramid based depth inference for multi-view stereo,'' in \emph{Proc. IEEE/CVF Conf. Comput. Vis. Pattern Recognit. (CVPR)}, 2020, pp. 4876--4885.

\bibitem{PatchMatchNet}
F.~Wang, S.~Galliani, C.~Vogel, P.~Speciale, and M.~Pollefeys, ``Patchmatchnet: Learned multi-view patchmatch stereo,'' in \emph{Proc. IEEE/CVF Conf. Comput. Vis. Pattern Recognit. (CVPR)}, 2021, pp. 14\,194--14\,203.

\bibitem{MVSTER}
X.~Wang, Z.~Zhu, G.~Huang, F.~Qin, Y.~Ye, Y.~He, X.~Chi, and X.~Wang, ``{{MVSTER}}: {{Epipolar}} transformer for efficient multi-view stereo,'' in \emph{European {{Conference}} on {{Computer Vision}}}.\hskip 1em plus 0.5em minus 0.4em\relax Springer, 2022, pp. 573--591.

\bibitem{EPP-MVSNet}
X.~Ma, Y.~Gong, Q.~Wang, J.~Huang, L.~Chen, and F.~Yu, ``Epp-mvsnet: Epipolar-assembling based depth prediction for multi-view stereo,'' in \emph{Proc. IEEE/CVF Int. Conf. Comput. Vis. (ICCV)}, 2021, pp. 5712--5720.

\bibitem{Geo-MVSNet}
Z.~Zhang, R.~Peng, Y.~Hu, and R.~Wang, ``{{GeoMVSNet}}: {{Learning Multi-View Stereo With Geometry Perception}},'' in \emph{Proceedings of the {{IEEE}}/{{CVF Conference}} on {{Computer Vision}} and {{Pattern Recognition}}}, 2023, pp. 21\,508--21\,518.

\bibitem{RA-MVSNet}
Y.~Zhang, J.~Zhu, and L.~Lin, ``Multi-{{View Stereo Representation Revist}}: {{Region-Aware MVSNet}},'' in \emph{Proceedings of the {{IEEE}}/{{CVF Conference}} on {{Computer Vision}} and {{Pattern Recognition}}}, 2023, pp. 17\,376--17\,385.

\bibitem{COLMAP}
J.~L. Sch{\"o}nberger, E.~Zheng, J.-M. Frahm, and M.~Pollefeys, ``Pixelwise view selection for unstructured multi-view stereo,'' in \emph{Proc. Eur. Conf. Comput. Vis. (ECCV)}, 2016, pp. 501--518.

\bibitem{Accurate}
S.~Shen, ``Accurate multiple view 3d reconstruction using patch-based stereo for large-scale scenes,'' \emph{IEEE Trans. Image Process.}, vol.~22, no.~5, pp. 1901--1914, 2013.

\bibitem{DepthPro}
A.~Bochkovskii, A.~Delaunoy, H.~Germain, M.~Santos, Y.~Zhou, S.~R. Richter, and V.~Koltun, ``Depth {{Pro}}: {{Sharp Monocular Metric Depth}} in {{Less Than}} a {{Second}},'' Oct. 2024.

\bibitem{Marigold}
B.~Ke, A.~Obukhov, S.~Huang, N.~Metzger, R.~C. Daudt, and K.~Schindler, ``Repurposing diffusion-based image generators for monocular depth estimation,'' in \emph{{IEEE/CVF} Conference on Computer Vision and Pattern Recognition, n {CVPR} 2024, Seattle, WA, USA, June 16-22, 2024}.\hskip 1em plus 0.5em minus 0.4em\relax {IEEE}, 2024, pp. 9492--9502.

\bibitem{NDDepth}
S.~Shao, Z.~Pei, W.~Chen, P.~C.~Y. Chen, and Z.~Li, ``Nddepth: Normal-distance assisted monocular depth estimation and completion,'' \emph{{IEEE} Trans. Pattern Anal. Mach. Intell.}, vol.~46, no.~12, pp. 8883--8899, 2024.

\bibitem{EP-Net}
W.~Su and W.~Tao, ``Efficient edge-preserving multi-view stereo network for depth estimation,'' in \emph{Proceedings of the {{AAAI Conference}} on {{Artificial Intelligence}}}, vol.~37, 2023, pp. 2348--2356.

\bibitem{TAPA-MVS}
A.~Romanoni and M.~Matteucci, ``Tapa-mvs: Textureless-aware patchmatch multi-view stereo,'' in \emph{Proc. IEEE/CVF Int. Conf. Comput. Vis. (ICCV)}, 2019.

\bibitem{ACMP}
Q.~Xu and W.~Tao, ``Planar {Prior} {Assisted} {PatchMatch} {Multi}-{View} {Stereo},'' in \emph{Proc. of the {AAAI} {Conf}. {Artif}. {Intell}. ({AAAI})}, April 2020.

\bibitem{OpenMVS}
\BIBentryALTinterwordspacing
D.~Cernea, ``{{OpenMVS}}: {{Multi-View Stereo Reconstruction Library}},'' 2020. [Online]. Available: \url{https://cdcseacave.github.io/openMVS}
\BIBentrySTDinterwordspacing

\bibitem{AGG-CVCNet}
L.~Xu, T.~Guan, Y.~Wang, Y.~Luo, Z.~Chen, W.~Liu, and W.~Yang, ``Self-supervised multi-view stereo via adjacent geometry guided volume completion,'' in \emph{{MM} '22: The 30th {ACM} International Conference on Multimedia, Lisboa, Portugal, October 10 - 14, 2022}, 2022, pp. 2202--2210.

\bibitem{PCF-MVS}
A.~Kuhn, S.~Lin, and O.~Erdler, ``Plane completion and filtering for multi-view stereo reconstruction,'' in \emph{Pattern Recognition}, 2019, pp. 18--32.

\end{thebibliography}
\bibliographystyle{IEEEtran}

\begin{IEEEbiography}[{\includegraphics[width=1in,height=1.3in,clip,keepaspectratio]{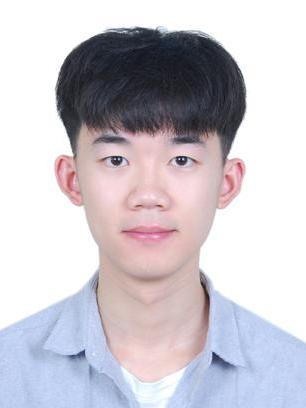}}]{Zhenlong Yuan}
received the B.Sc. degree in telecommunications engineering with management from Beijing University of Posts and Telecommunications, Beijing, China. He is working toward the Ph.D. degree in the Institute of Computing Technology, Chinese Academy of Sciences and University of Chinese Academy of Sciences. His main research interests include vision-language model and 3D reconstruction.
\end{IEEEbiography}

\begin{IEEEbiography}[{\includegraphics[width=1in,height=1.3in,clip,keepaspectratio]{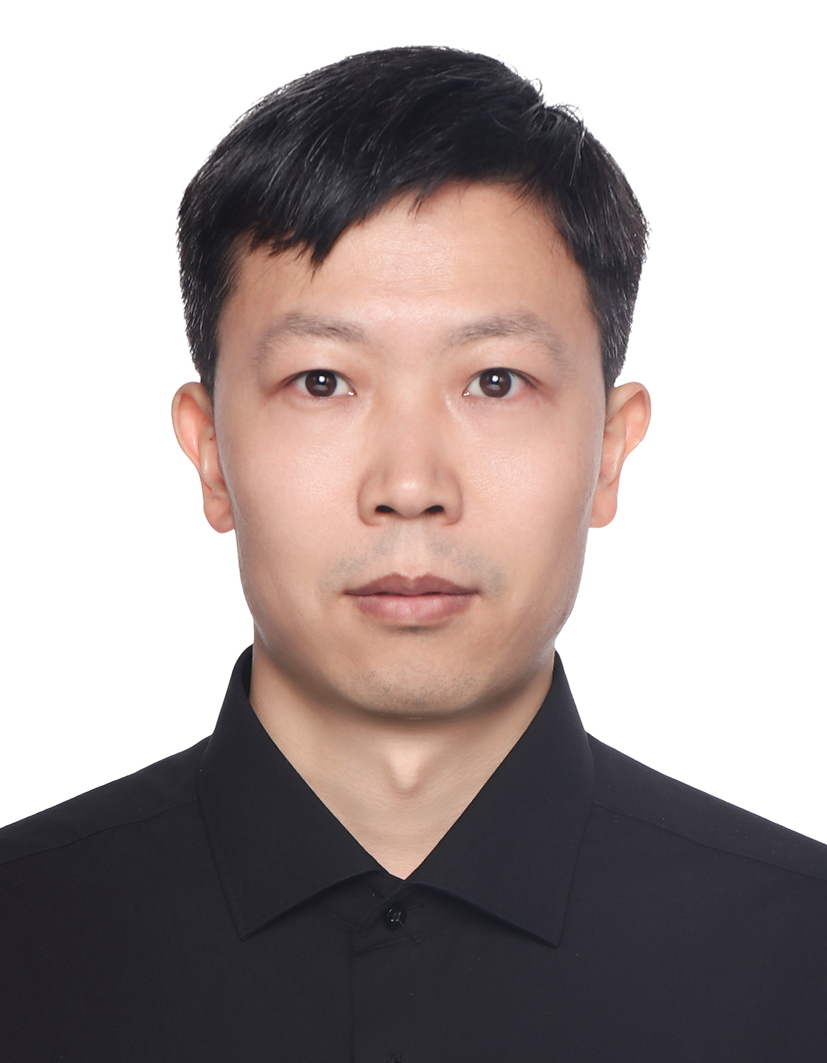}}]{Dapeng Zhang}
is a Ph.D. candidate in the School of Information Science and Engineering at Lanzhou University. His research focuses on autonomous driving and world model.
\end{IEEEbiography}

\begin{IEEEbiography}[{\includegraphics[width=1in,height=1.3in,clip,keepaspectratio]{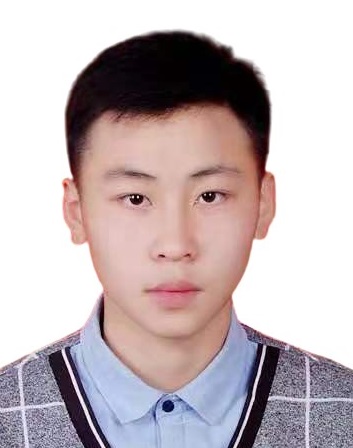}}]{Zehao Li}
received the B.Sc. degree in computer science and technology from Yunnan University, Kunming, China, in 2023. He is currently pursuing the Ph.D. degree with the Institute of Computing Technology, Chinese Academy of Sciences, Beijing, China. His main research interests include computer graphics and computer vision.
\end{IEEEbiography}

\begin{IEEEbiography}[{\includegraphics[width=1in,height=1.3in,clip,keepaspectratio]{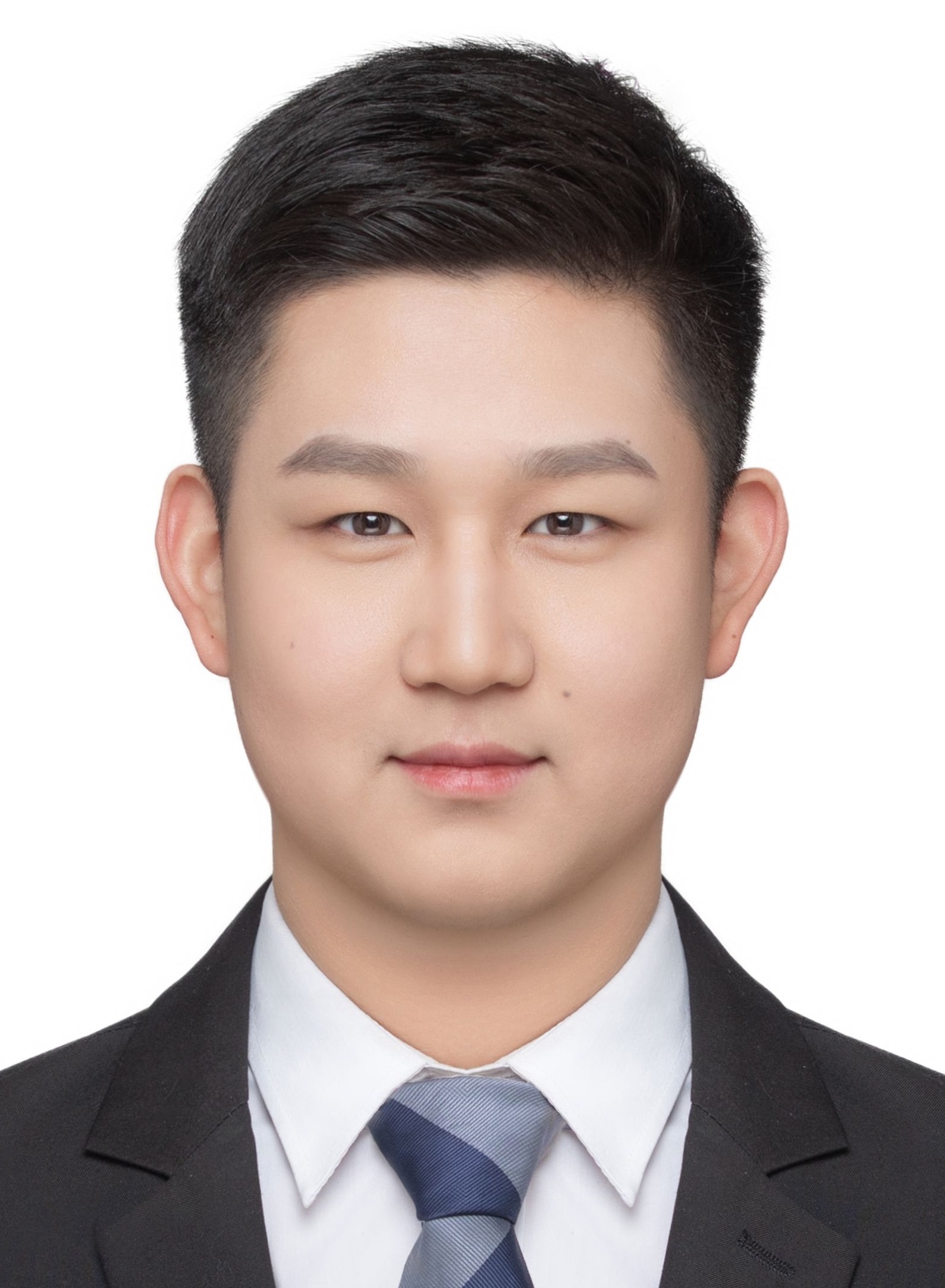}}]{Chengxuan Qian} 
is currently pursuing a B.S. degree in Mathematics and Applied Mathematics at the School of Mathematical Sciences, Jiangsu University, from 2022 to 2026. His research interests include Multimodal Large Language Models, 3D Vision, Embodied AI and Multimodal Representation Learning. 
\end{IEEEbiography}

\begin{IEEEbiography}[{\includegraphics[width=1in,height=1.3in,clip,keepaspectratio]{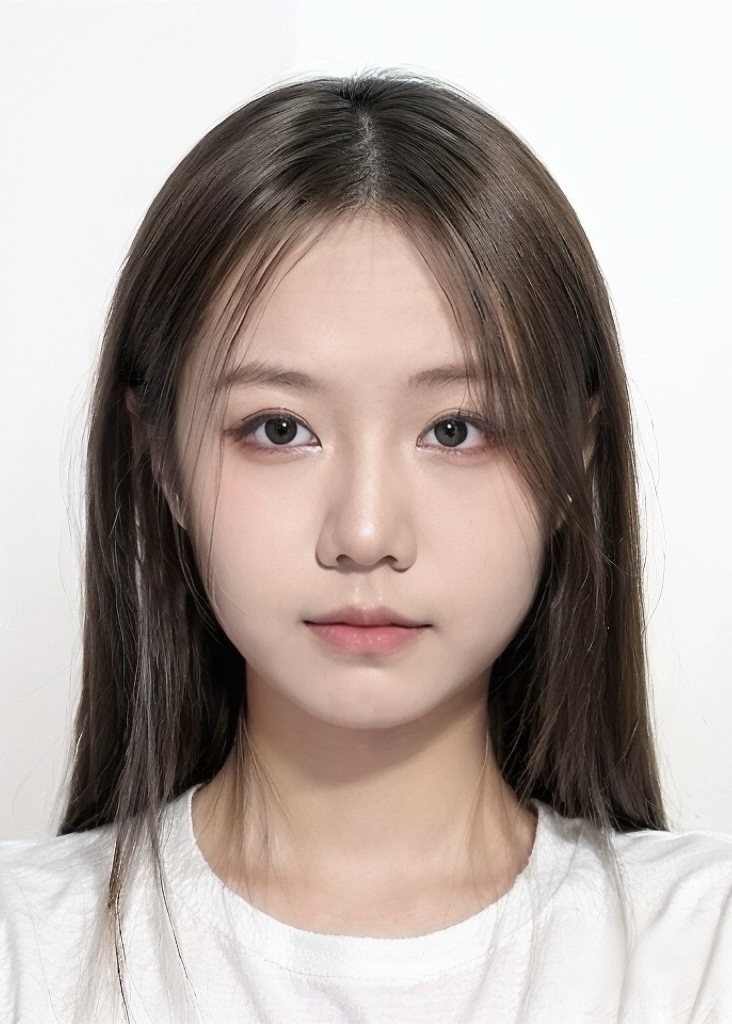}}]{Jianing Chen}
received B.E. degree in software engineering from Chinese University of Geoscience (Beijing), Beijing, China, in 2023. She is currently pursuing an M.S. degree in computer science and technology at the Institute of Computer Technology, Chinese Academy of Science. Her research interests include computer vision and 3D reconstruction.
\end{IEEEbiography}

\begin{IEEEbiography}[{\includegraphics[width=1in,height=1.3in,clip,keepaspectratio]{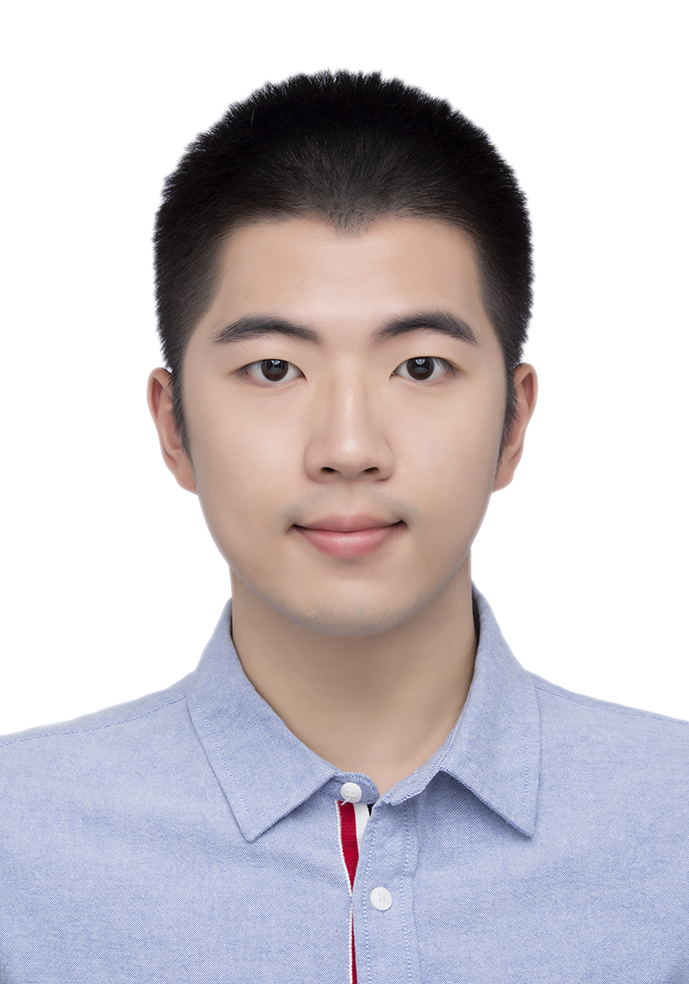}}]{Yinda Chen}
received the dual B.S. degrees in Remote Sensing and Economics from Xiamen University. He is currently pursuing the Ph.D. degree in Information and Communication Technology at the University of Science and Technology of China. His research interests include self-supervised pretraining and image compression.
\end{IEEEbiography}

\begin{IEEEbiography}[{\includegraphics[width=1in,height=1.3in,clip,keepaspectratio]{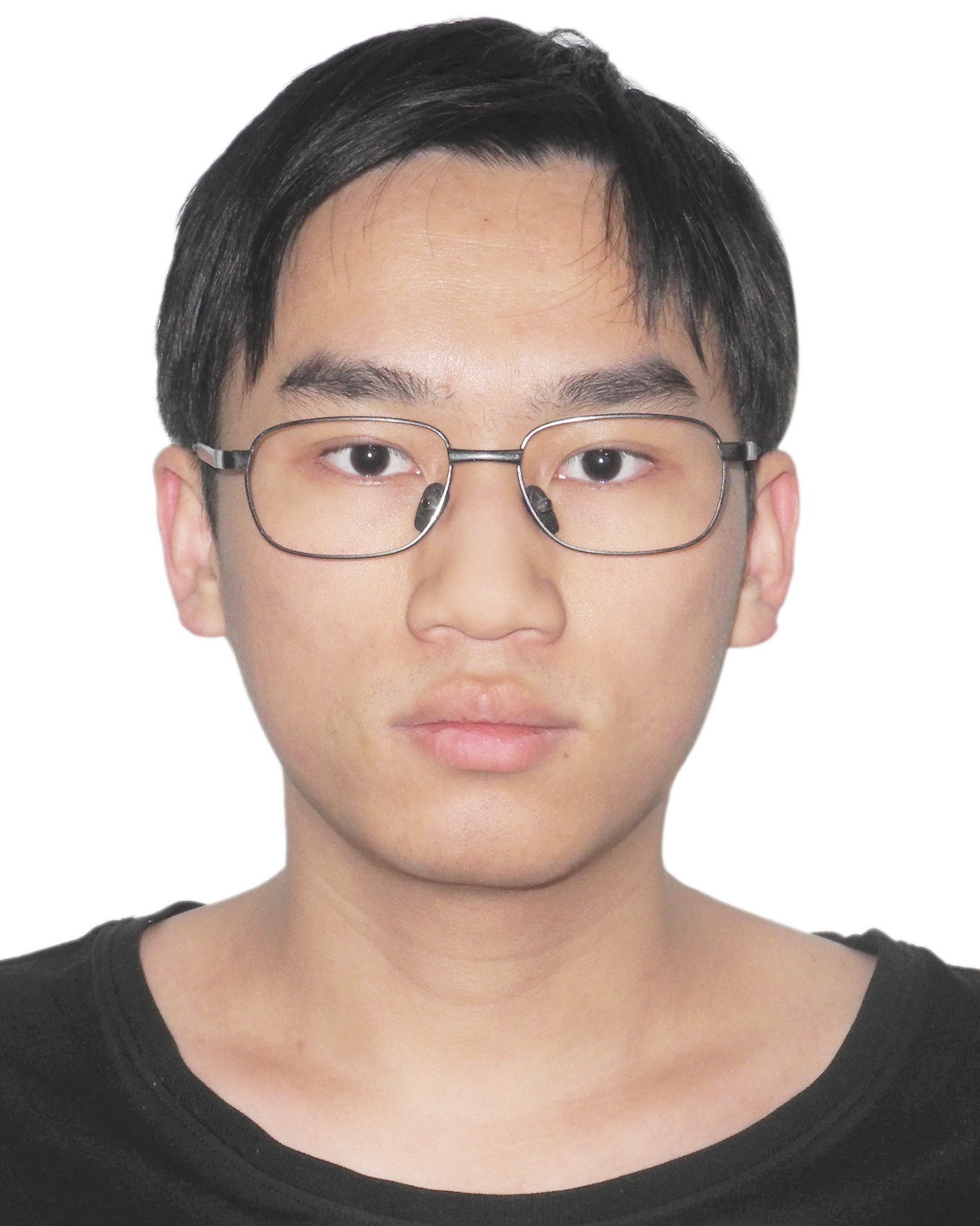}}]{Kehua Chen}
received his Bachelor’s degree in Software Engineering from Jilin University. He is currently pursuing his Master’s degree at the Institute of Computing Technology, Chinese Academy of Sciences. His main research interests include multiview stereo and 3D vision.
\end{IEEEbiography}

\begin{IEEEbiography}[{\includegraphics[width=1in,height=1.3in,clip,keepaspectratio]{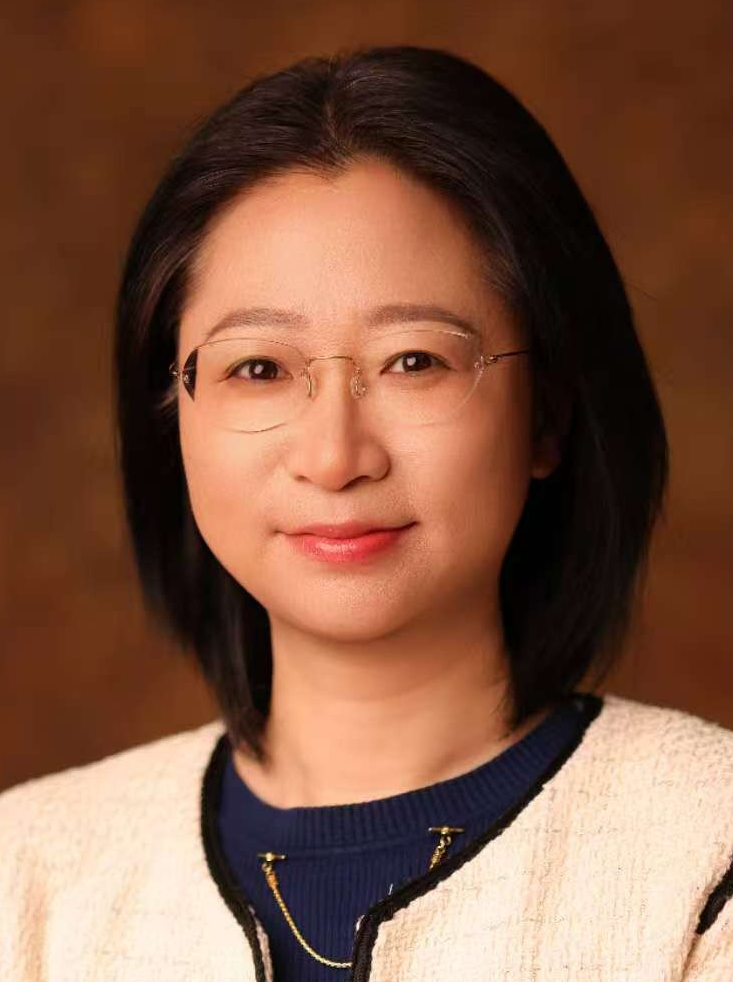}}]{Tianlu Mao}
received the Ph.D. degree from the Institute of Computing Technology, Chinese Academy of Sciences, in 2009, where she is also working as associate professor. Her research in-terests include artificial intelligence,  modeling and simulation.
\end{IEEEbiography}

\begin{IEEEbiography}[{\includegraphics[width=1in,height=1.3in,clip,keepaspectratio]{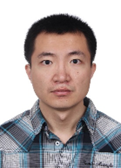}}]{Zhaoxin Li}
received the Ph.D. degree in computer application technology from Harbin Institute of Technology, Harbin, China, in 2016. From July 2016 to April 2023, he worked as an Assistant Professor in the Institute of Computing Technology, Chinese Academy of Sciences. From September 2018 to March 2019, he worked as a Postdoctoral Fellow in the Department of Computing, The Hong Kong Polytechnic University. He is currently with the Agricultural Information Institute of Chinese Academy of Agricultural Sciences, China. His research interests include 3D computer vision and 3D data processing.
\end{IEEEbiography}

\begin{IEEEbiography}[{\includegraphics[width=1in,height=1.3in,clip,keepaspectratio]{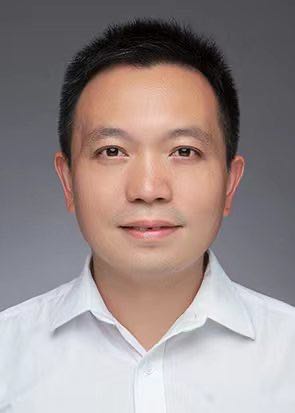}}]{Hao Jiang}
obtained Ph.D from the Institute of Computing Technology, Chinese Academy of Sciences. Currently, he is an associate professor at the Institute of Computing Technology, Chinese Academy of Sciences and University of Chinese Academy of Sciences. His research interests include virtual reality/augmented reality, computer graphics, and intelligent user interfaces.
\end{IEEEbiography}

\begin{IEEEbiography}[{\includegraphics[width=1in,height=1.3in,clip,keepaspectratio]{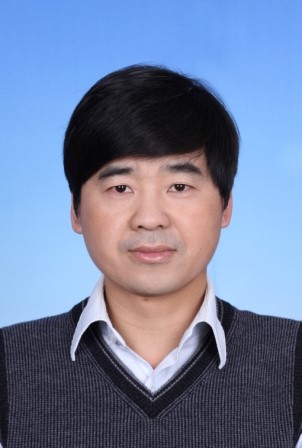}}]{Zhaoqi Wang}
is a researcher and a director of PhD students with the Institute of Computing Technology, Chinese Academy of Sciences. His research interests include virtual reality and intel-ligent human computer interaction. He is a senior member of the China Computer Federation.
\end{IEEEbiography}

\vfill

\end{document}